\documentclass[10pt,twocolumn,letterpaper]{article}
\usepackage[table,dvipsnames]{xcolor}
\usepackage{tikz}  %
\usepackage{cvpr}
\usepackage{times}
\usepackage{graphicx}
\usepackage{amsmath}
\usepackage{amssymb}
\usepackage{textcomp}
\usepackage{makecell}

\usepackage[caption=false]{subfig}

\usepackage{subfiles}
\usepackage{pifont} %
\usepackage{pgfplots} %
\usepackage{pgfplotstable} %
\pgfplotsset{compat=newest} %
\usetikzlibrary{plotmarks} %
\usetikzlibrary{colorbrewer} %
\usepackage{booktabs} %
\usepackage{multirow}
\usepackage{etoolbox,siunitx} %

\robustify\bfseries
\robustify\itshape
\usepackage{paralist} %
\newcommand{\reffig}[1]{Fig.~\ref{#1}}

\definecolor{olivegreen}{RGB}{0,170,0}
\definecolor{darkred}{RGB}{220,100,10}
\definecolor{tealblue}{RGB}{20,100,200}

\definecolor{rowblue}{RGB}{220,230,240}
\definecolor{rowcyan}{RGB}{220,240,240}

\definecolor{citecolor}{RGB}{34,139,34}

\newcommand{\feat}{\mathrm{F}}
\newcommand{\set}[1]{\mathcal{S}_{#1}}
\newcommand{\mycomment}[1]{}

\newcommand{\tablestyle}[2]{\setlength{\tabcolsep}{#1}\renewcommand{\arraystretch}{#2}\centering\footnotesize}
\newlength\savewidth\newcommand\shline{\noalign{\global\savewidth\arrayrulewidth
  \global\arrayrulewidth 1pt}\hline\noalign{\global\arrayrulewidth\savewidth}}
\newcolumntype{x}[1]{>{\centering\arraybackslash}p{#1pt}}
\newcommand{\bftab}{\fontseries{b}\selectfont}

\newcommand{\app}{\raise.17ex\hbox{$\scriptstyle\sim$}}

\newcommand{\showonerowpascal}[7]{%
\setlength{\fboxsep}{0pt}
\resizebox{#6\linewidth}{!}{%
    {#7 \rotatebox{90}{\centering \hspace{#5mm}\vphantom{tg}#4}}
	\fbox{\includegraphics[height=#3\linewidth]{ims/#2/#1.jpg}}\hspace{2mm}
    \hfill \fbox{\includegraphics[height=#3\linewidth]{ims/#2/#1_edge.png}}\hspace{2mm}
	\hfill \fbox{\includegraphics[height=#3\linewidth]{ims/#2/#1_semseg.png}}\hspace{2mm}
	\hfill \fbox{\includegraphics[height=#3\linewidth]{ims/#2/#1_human_parts.png}}\hspace{2mm}
	\hfill \fbox{\includegraphics[height=#3\linewidth]{ims/#2/#1_normals.png}}\hspace{2mm}
    \hfill \fbox{\includegraphics[height=#3\linewidth]{ims/#2/#1_sal.png}}
	}\\[1mm]
}

\newcommand{\showonerownyud}[6]{%
\setlength{\fboxsep}{0pt}
\resizebox{#6\linewidth}{!}{%
    {\huge \rotatebox{90}{\centering \hspace{#5mm}\vphantom{tg}#4}}
	\fbox{\includegraphics[height=#3\linewidth]{ims/#2/#1.jpg}}\hspace{2mm}
	\hfill
	\fbox{\includegraphics[height=#3\linewidth]{ims/#2/#1_edge.png}}\hspace{2mm}
	\hfill
	\fbox{\includegraphics[height=#3\linewidth]{ims/#2/#1_semseg.png}}\hspace{2mm}
	\hfill
	\fbox{\includegraphics[height=#3\linewidth]{ims/#2/#1_normals.png}}\hspace{2mm}
	\hfill
	\fbox{\includegraphics[height=#3\linewidth]{ims/#2/#1_depth.png}}
	}\\[1mm]
}

\usepackage[pagebackref=true,breaklinks=true,letterpaper=true,colorlinks,citecolor=citecolor,bookmarks=false]{hyperref}

\cvprfinalcopy %

\def\cvprPaperID{651} %

\ifcvprfinal\fi
\begin{document}
\title{Attentive Single-Tasking of Multiple Tasks}

\author{Kevis-Kokitsi Maninis\thanks{Work done while at Facebook AI Research.}\\Computer Vision Lab, ETH Z\"urich \and Ilija Radosavovic\\Facebook AI Research (FAIR) \and Iasonas Kokkinos$^{*}$\\Ariel AI, UCL
}

\maketitle

\begin{abstract}
In this work we address task interference in universal networks by considering that a network is trained on multiple tasks, but performs one task at a time, an approach we refer to as ``single-tasking multiple tasks''.
The network thus modifies its behaviour through task-dependent feature adaptation, or task attention. 
This gives the network the ability to accentuate the features that are adapted to a task, while shunning irrelevant ones. 
We further reduce task interference by forcing the task gradients to be statistically indistinguishable through adversarial training,  
ensuring that the common backbone architecture serving all tasks is not dominated by any of the task-specific gradients.   

Results in three multi-task dense labelling problems consistently show: (i)  a large reduction in the number of parameters while
preserving, or even improving performance and (ii) a smooth trade-off between computation and multi-task accuracy. We provide our system's code and pre-trained models at \url{http://vision.ee.ethz.ch/~kmaninis/astmt/}.
\end{abstract}
\vspace{-5mm}
	
\section{Introduction}
Real-world  problems involve a multitude of visual tasks that call for multi-tasking, universal vision systems. For instance autonomous driving requires detecting pedestrians, estimating velocities and reading traffic signs, while identity recognition, pose, face  and hand tracking are required for human-computer interaction. 

A thread of works have introduced multi-task networks \cite{Ser+14,EiFe15,He+17,Kokk17} handling an increasingly large number of tasks. 
Still, it is common practice to train devoted networks for individual tasks when single-task performance is critical.
This is supported by negative results from recent works that have aimed at addressing multiple problems with a single  network \cite{He+17,Kokk17} - these have shown that there is a limit on performance imposed by the capacity of the network, manifested as a drop in performance when loading a single network with more tasks. Stronger backbones can uniformly improve multi-task performance, but
still the per-task performance can be lower than the single-task performance with the same backbone.

\begin{figure}[t]
\centering
\resizebox{.99\linewidth}{!}{%
\includegraphics[width=1\linewidth]{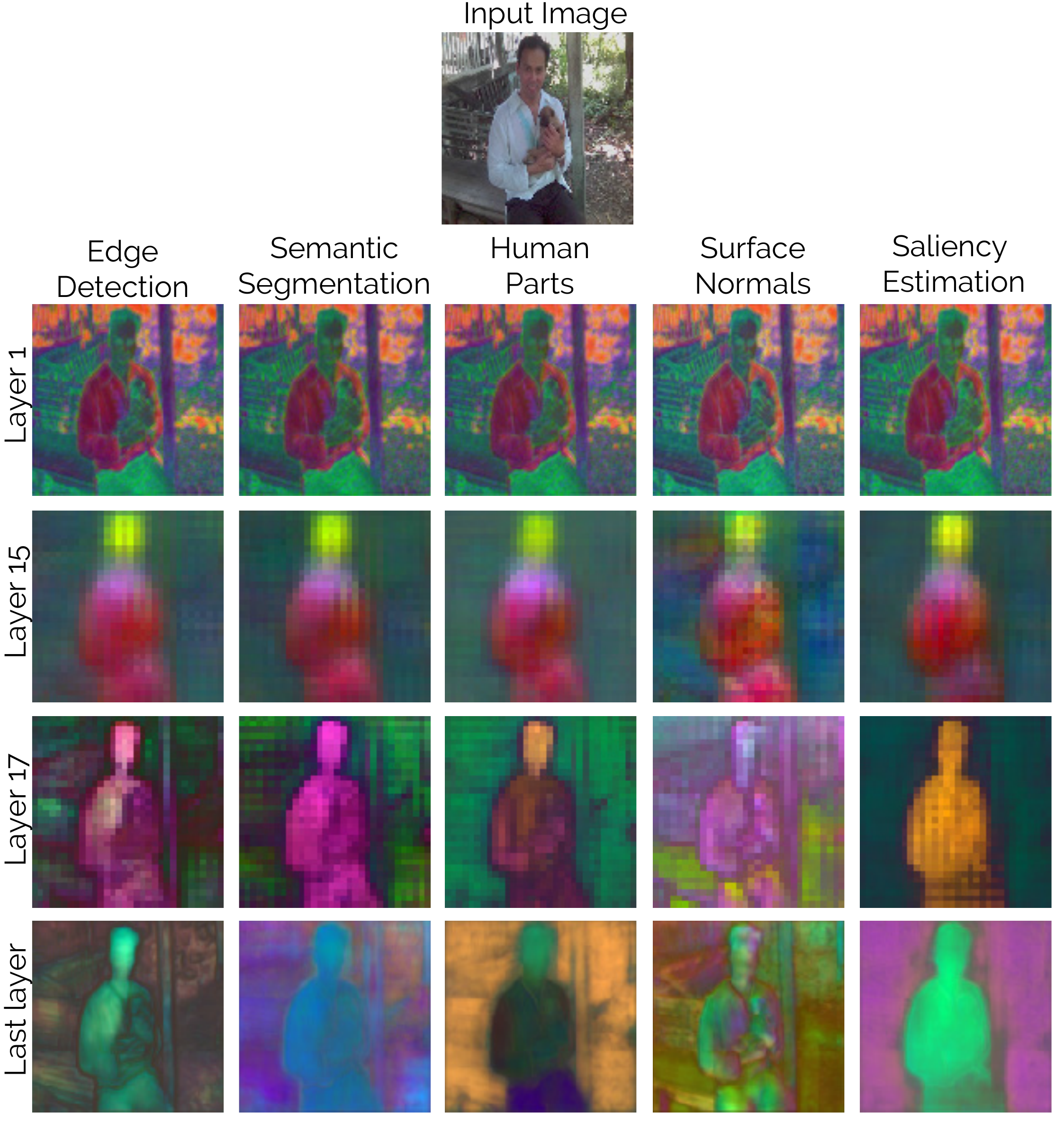}
}
\caption{\textbf{Learned representations across tasks and  layers:} We visualize how features change spatially in different depths of our multi-task network. For each layer (row) we compute a common PCA basis across tasks (column) and show the first three principal components as RGB values at each spatial location. We observe that the features are more similar in early layers and get more adapted to specific tasks as depth increases, leading to disentangled, task-specific representations in the later layers. We see for instance that the normal task features co-vary with surface properties, while the part features remain constant in each human part.}
\label{fig:spat_pca_3_0}
\vspace{-1em}
\end{figure}

\begin{figure}[t]
\centering
\resizebox{1\linewidth}{!}{%
\includegraphics[width=1\linewidth]{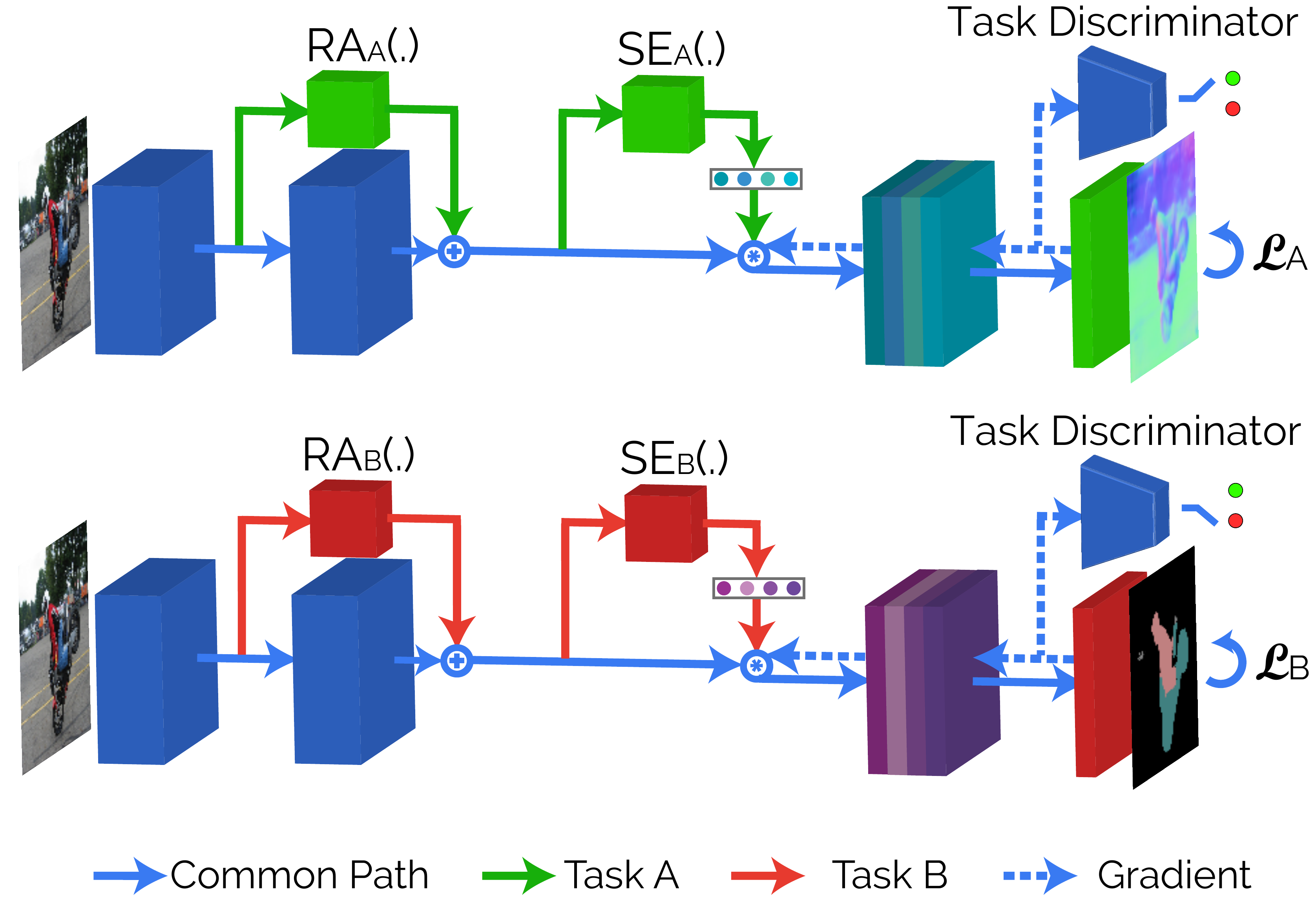}
}
\caption{While using a shared backbone network, every task adapts its behavior in a separate, flexible, and lightweight manner, allowing us to customize computation for the task at hand. 
We refine features with a task-specific residual adapter branch (RA), and attend to particular channels with task-specific Squeeze-and-Excitation (SE) modulation. We also enforce the task gradients (dashed lines) to be statistically indistinguishable through adversarial training, further promoting the separation between task-specific and generic layers. 
}
\label{fig:cover}
\vspace{-1em}
\end{figure}

This problem, known as task interference, can be understood as facing a the dilemma of invariance versus sensitivity: the most crucial information for one task can be a nuisance parameter for another, which leads to potentially conflicting objectives when training multi-task networks.
An example of such a task pair is pose estimation and object detection: when detecting or counting people the detailed pose information is a nuisance parameter that should be eliminated at some point from the representation of a network aiming at pose invariance~\cite{He+17}.
At the same time, when watching a dance performance, one needs the detailed pose of the dancers, while ignoring the large majority of spectators.
More generally this is observed when combining a task that is detail-oriented and requires high spatial acuity with a task that requires abstraction from spatial details, e.g. when one wants to  jointly do low- and high-level vision. In other words, one task's noise is another one's signal. 

We argue that this dilemma can be addressed by single-tasking, namely executing task a time, rather than getting all task responses in a single forward pass through the network. This reflects many practical setups, for instance when one sees the results of a single computational photography task at a time on the screen of a mobile phone, rather than all of them jointly. 
Operating in this setup allows us to implement an  ``attention to task'' mechanism that  changes the network's behaviour in a task-adapted manner, as shown in \reffig{fig:spat_pca_3_0}. 
We use the exact same network backbone in all cases, but we modify the network's behavior according to the executed task by relying on the most task-appropriate features.
For instance when performing a low-level task such as boundary detection or normal estimation, the network can retain and elaborate on fine image structures, while shunning them for a high-level task that requires spatial abstraction. 

We explore two different {\emph{task attention mechanisms}}, as shown in \reffig{fig:cover}.
Firstly, we  use  data-dependent modulation signals~\cite{Per+17} that enhance or suppress neuronal activity in a task-specific manner.
Secondly, we use  task-specific Residual Adapter~\cite{RBV18} blocks that latch on to a larger architecture in order to extract task-specific information which is fused with the  representations extracted by a generic backbone.
This allows us to learn a shared backbone representation that serves all tasks but collaborates with task-specific processing to build more elaborate task-specific features. 

These two extensions can be understood as promoting a disentanglement between the shared representation learned across all tasks and the task-specific parts of the network. Still, if the loss of a single task is substantially larger, its gradients will overwhelm those of others and disrupt the training of the shared representation. In order to make sure that no task abuses the shared resources we impose a {\em{task-adversarial loss}} to the network gradients, requiring that these are statistically indistinguishable across tasks. This loss is minimized during training through double back-propagation \cite{DrLe91}, and leads to an automatic balancing of loss terms, while promoting compartmentalization between task-specific and shared blocks.

\section{Related Work}

Our work draws ideas from several research threads. %
\textbf{Multiple Task Learning (MTL):} Several works have shown that jointly learning pairs of tasks yields fruitful results in computer vision. Successful pairs include detection and classification~\cite{Girs15,Ren+15}, detection and segmentation~\cite{He+17, Dvo+17}, or monocular depth and segmentation~\cite{EiFe15,Xu+18}. Joint learning is beneficial for unsupervised learning~\cite{Ran+18}, when tasks provide complementary information (eg. depth boundaries and motion boundaries~\cite{ZLH18}), in cases where task A acts as regularizer for task B due to limited data~\cite{LQH17}, or in order to learn more generic representations from synthetic data~\cite{ReLe18}. Xiao et al.~\cite{Xia+18} unify inhomogeneous datasets in order to train for multiple tasks, while~\cite{Zam+18} explore relationships among a large amount of tasks for transfer learning, reporting improvements when transferring across particular task pairs. 

Despite these  positive results, joint learning can be harmful in the absence of a direct relationship between  task pairs. This was reported clearly in \cite{Kokk17} where the joint learning of low-, mid- and high-level tasks was explored, reporting that the improvement of one task (e.g. normal detection) was to the detriment of another (e.g. object detection). Similarly, when jointly training for human pose estimation on top of detection and segmentation, Mask R-CNN performs worse than its two-task counterpart~\cite{He+17}. 

This negative result first requires carefully calibrating the relative losses of the different tasks, so that none of them deteriorates excessively.  To address this problem, GradNorm~\cite{Che+17a} provides a method to adapt the weights such that each task contributes in a balanced way to the loss, by normalizing the gradients of their losses;  a more recent work~\cite{Sin+18} extends this approach to homogenize the task gradients based on adversarial training. Following a probabilistic treatment~\cite{KGC18} re-weigh the losses according to each task's uncertainty, while Sener and Koltun~\cite{SeKo18} estimate an adaptive weighting of the different task losses based on a pareto-optimal formulation of MTL. Similarly,~\cite{Guo+18} provide a MTL framework where tasks are dynamically sorted by difficulty and the hardest are learned first. 

A second approach to mitigate task interference consists in avoiding the `spillover' of gradients from one task's loss to the common features serving all tasks. One way of doing this is explicitly constructing complementary task-specific feature representations~\cite{PNN,RoTs18}, but results in an increase of complexity that is linear in the number of tasks. An alternative, adopted in the related problem of lifelong learning consists in removing from the gradient of a task's loss those components that would incur an increase in the loss of previous tasks~\cite{EWC,GEM}. For domain adaptation~\cite{Bou+16} disentangle the representations learned by shared/task-specific parts of networks by enforcing similarity/orthogonality constraints. Adversarial Training has been used in the context of domain adaptation~\cite{GaLe15, LQH17} to the feature space in order to fool the discriminator about the source domain of the features. 

In our understanding these losses promote a compartmental operation of a network, achieved  for instance when a  block-structured weight matrix  prevents the interference of features for tasks that should not be connected. A deep single-task implementation of this would be the gating mechanism of~\cite{AhTo18}. For multi-tasking, Cross Stitch Networks~\cite{Mis+16} automatically learn to split/fuse two independent networks in different depths according to their learned tasks, while~\cite{Mur+16} estimate a block-structured weight matrix during CNN training.

As we show in the next section, a soft ``compartmentalization'' effect that does not interfere with the network's weight matrix can be achieved in the case of multi-task learning  through a soft, learnable form of task-specific feature masking. This shares several similarities with attention mechanisms, briefly described below.

\textbf{Attention mechanisms:}
Attention has often been used in deep learning to visualize and interpret the inner workings of CNNs~\cite{SVZ13,ZeFe14,Sel+17}, but has also  been employed to improve the learned representations of convolutional networks for scale-aware semantic segmentation~\cite{Che+17}, fine-grained image recognition~\cite{FZM17} or caption generation~\cite{Xu+15,Lu+16,And+18}. Squeeze and Excitation Networks~\cite{HSS18} and their variants~\cite{Woo+18,Hu+18a} modulate the information of intermediate spatial features according to a global representation and be understood as implementing attention to different channels. Deep Residual Adapters~\cite{RBV17,RBV18} modulate learned representations depending on their source domain. Several works study modulation for image retrieval~\cite{Zha+18} or classification tasks~\cite{Per+17, Mud+19}, and embeddings for different artistic styles~\cite{DSK17}. \cite{Yan+18} learns object-specific modulation signals for video object segmentation, and~\cite{RBT18} modulates features according to given priors for detection and segmentation. In our case we learn task-specific modulation functions that allow us to drastically change the  network's behaviour while using identical backbone weights.

\section{Attentive Single-Tasking Mechanisms}
\label{sec:modulation}

\begin{figure}[t]
\centering
\includegraphics[height=.3\linewidth]{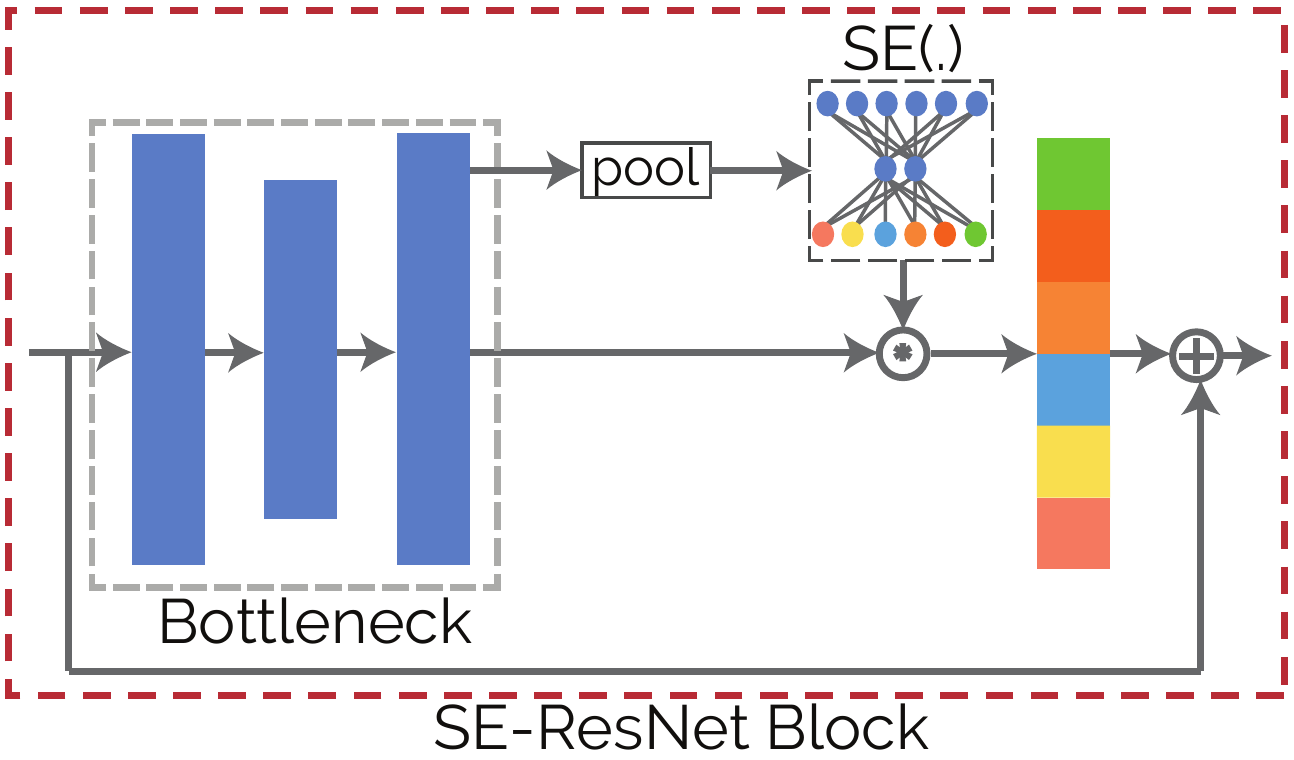} \\
\includegraphics[height=.4\linewidth]{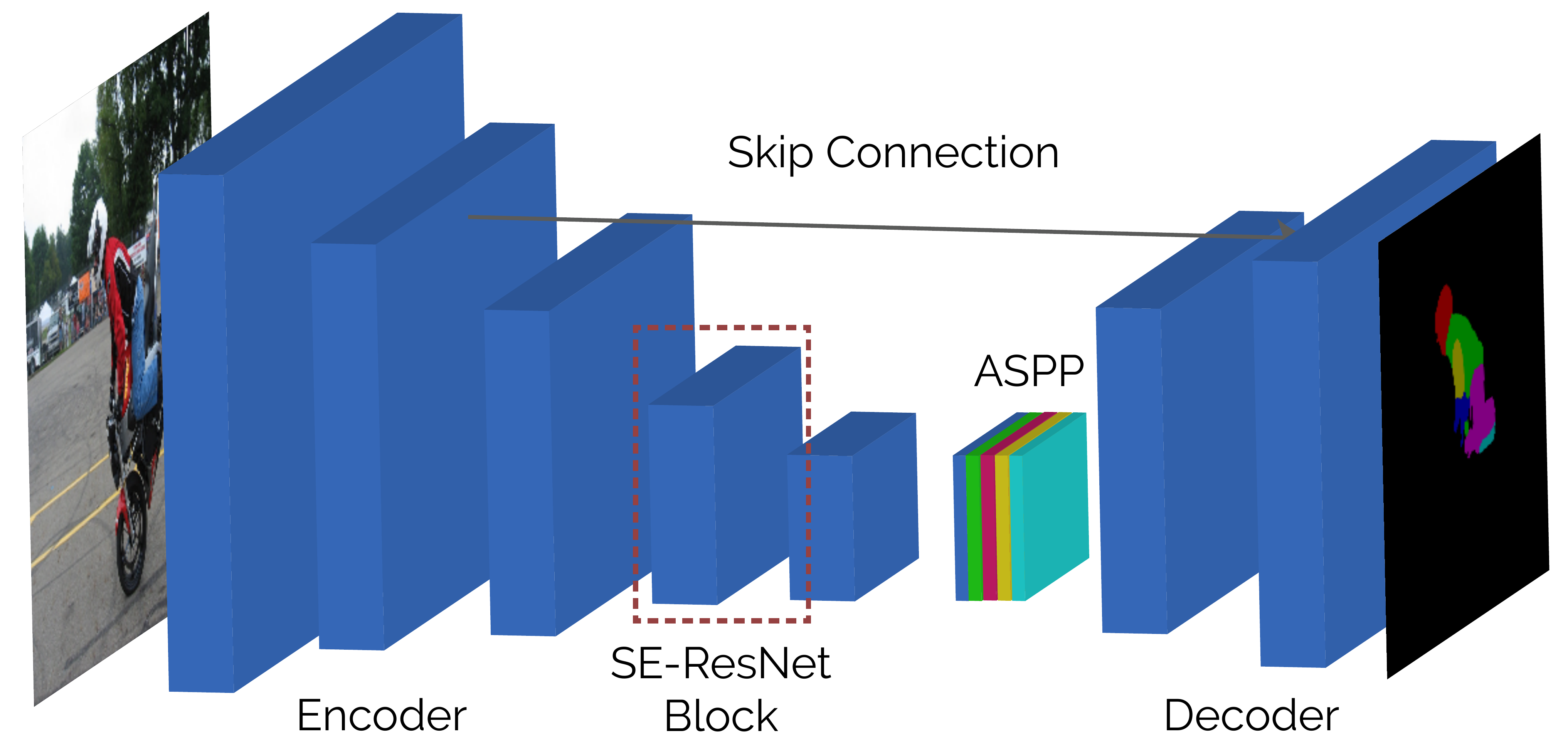}

\caption{\textbf{Single-task network architecture:} We use Deeplab-v3+ with a Squeeze-and-Excitation (SE)-ResNet backbone. SE modules are present in all bottleneck blocks of the encoder and the decoder. Attentive multi-tasking uses  different SE layers per task to modulate the network features in a task-specific manner.}
\label{fig:arch_se}
\vspace{-.5em}
\end{figure}

Having a shared representation for multiple tasks
can be efficient from the standpoint of memory- and sample- complexity, but can result in worse performance if the same resources are serving tasks with unrelated, or even conflicting objectives, as described above. 
Our proposed remedy to this problem consists in learning a shared representation for all tasks, while allowing each task to use this shared representation differently for the construction of its own features. \\

\subsection{Task-specific feature modulation}
In order to justify  our approach we start with a minimal example. 
We consider that we have two tasks A and B that share a common feature  tensor $\feat(x,y,c)$ at a given network layer, where $x,y$ are spatial coordinates and $c=1,\ldots,C$ are the tensor channels. 
We further assume that a subset $\set{A}$ of the channels is better suited for task A, while $\set{B}$ is better for B. 
For instance if A is invariant to deformations (detection) while B is sensitive (pose estimation), 
$\set{A}$ could be features obtained by taking more context into account, while $\set{B}$ would be more localized. 

One simple  way of ensuring that tasks A and B do not interfere  while using a shared feature tensor is to hide the features of task B when training for task A:
\begin{equation}
\feat_{A}(x,y,c) = m_A[c] \cdot \feat(x,y,c) 
\end{equation}
where $m_A[c] \in \{0,1\}$ is the indicator function  of set $\set{A}$.
If $c \notin \set{A}$ then $\feat_{A}(x,y,c)=0$, which means that the gradient $\frac{\partial \mathrm{\mathcal{L}_A}}{\partial \feat(x,y,c)}$ sent by the loss $\mathcal{L}_A$ of task A to $c \in \set{A}$ will be zero.
We thereby avoid task interference since Task A does not influence nor use features that it does not need. 

Instead of this hard choice of features per task we opt for a soft, differentiable membership function that is learned in tandem with the network and allows the different tasks to discover during training which features to use. Instead of a constant  membership function per channel we opt for an image-adaptive term that allows one to exploit the power of the squeeze-and-excitation block~\cite{HSS18}.

In particular we adopt the squeeze-and-excitation (SE) block (also shown in Fig.~\ref{fig:cover}), combining a global average pooling operation of the previous layer with a fully-connected layer that feeds into a sigmoid function, yielding a differentiable, image-dependent channel gating function. We set the parameters of this layer to be task-dependent, allowing every task to modulate the available channels differently. 
As shown in Section~\ref{sec:experiments}, this can result in substantial improvements when compared to a baseline that uses the same SE block for all tasks. 
\subsection{Residual Adapters}
The feature modulation described above can be understood as shunning those features that do not contribute to the task while focusing on the more relevant ones. Intuitively, this does not add capacity to the network but rather cleans the signal that flows through it from information that the task should be invariant to. 
We propose to complement this by appending task-specific sub-networks that adapt and refine the shared features in terms of residual operations of the following form:
\begin{equation}
\mathrm{L}_{A}(x) = x + \mathrm{L}(x) + \mathrm{RA}_{A}(x),
\end{equation}
where $\mathrm{L}(x)$ denotes the default behaviour of a residual layer, $\mathrm{RA}_{A}$ is the task-specific residual adapter of task $A$, and $\mathrm{L}_{A}(x)$ is the modified layer. 
We note that if $\mathrm{L}(x)$ and $\mathrm{RA}_{A}(x)$ were linear layers this would amount to the classical regularized multi-task learning of \cite{EvPo04}.

These adapters introduce a task-specific parameter and computation budget that is used in tandem with that of the shared backbone network. We show in Section~\ref{sec:experiments} that this is typically a small fraction of the budget used for the shared network, but improves accuracy substantially.

When employing disentangled computation graphs with feature modulation through SE modules and/or residual adapters, we also use task-specific batch-normalization layers, that come with a trivial increase in parameters (while the computational cost remains the same).

\begin{figure}[t]
\centering
\resizebox{1\linewidth}{!}{%
\includegraphics[width=1\linewidth]{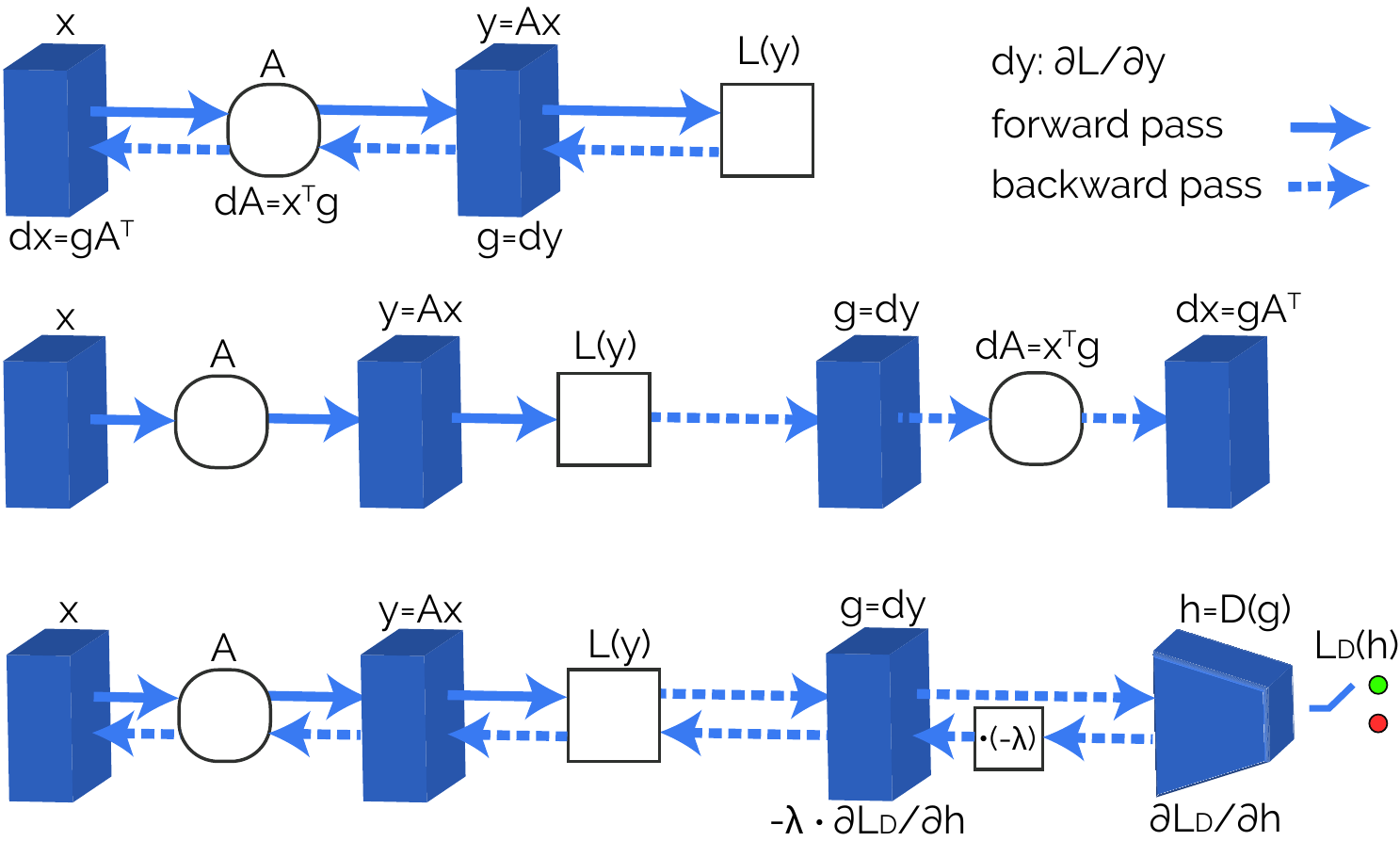}
}
\caption{Double backprop \cite{DrLe91}  exposes the gradients computed during backprop (row 1) by unfolding the computation graph of gradient computation (row 2). Exposing the gradients allows us to train them in an adversarial setting by using a discriminator, forcing them to be statistically indistinguishable across tasks (row 3). The shared network features $x$ then receive gradients that have the same distribution irrespective of the task, ensuring that no task abuses the shared network, e.g. due to higher loss magnitude. The gradient of the discriminator is reversed (negated) during adversarial training, and the parameter $\lambda \in [0, 1]$ controls the amount of negative gradient that flows back to the network~\cite{GaLe15}.
\label{fig:dscr}}
\vspace{-.5em}
\end{figure}

\section{Adversarial Task Disentanglement}
\label{sec:discriminator}

The idea behind the task-specific adaptation mechanisms described above is that even though a shared representation has better memory/computation complexity,  every task can profit by having its own `space', i.e. separate modelling capacity to make the best use of the representation - by modulating the features or adapting them with residual blocks.

Pushing this idea further we enforce a strict separation of the shared and task-specific processing, by requiring that the gradients used to train the shared parameters are statistically indistinguishable across tasks. This ensures that the shared backbone serves all tasks equally well, and is not disrupted e.g. by one task that has larger gradients. 

We enforce this constraint through  adversarial learning. Several methods, starting from Adversarial Domain Adaptation \cite{Gan+16}, use adversarial learning to remove any trace of a given domain from the learned mid-level features in a network; a technique called Adversarial multi-task training \cite{LQH17} falls in the same category. 

Instead of removing domain-specific information from the features of a network (which serves domain adaptation), we remove any task-specific trace from the gradients sent from different tasks to the shared backbone (which serves a division between shared and task-specific processing). A concurrent work~\cite{Sin+18} has independently proposed this idea.

\begin{table}
\rowcolors{2}{white}{white}
\resizebox{.5\textwidth}{!}{%
	\begin{tabular}{lc|cc|ccccccc}
		Database        & Type          & \# Train Im.   & \# Test Im.        & Edge	    & S.Seg      & H. Parts   & Normals        & Saliency       & Depth      & Albedo     \\
		\shline
		PASCAL			& Real	       	& 4,998          & 5,105 	   	      & \checkmark  & \checkmark & \checkmark & \checkmark$^*$ & \checkmark$^*$ &            &            \\
	    NYUD	        & Real			& 795		     & 654			      & \checkmark  & \checkmark & 	          & \checkmark     &                & \checkmark &            \\
		FSV				& Synthetic		& 223,197		 & 50,080 		 	  &             & \checkmark & 	          &                &                & \checkmark & \checkmark \\
	\end{tabular}
	}
\vspace{2mm}
\caption{\textbf{Multi-task benchmark statistics}: We conduct the main volume of experiments on PASCAL for 5 tasks ($^*$ labels obtained via distillation). We also use the fully labelled subsets of NYUD, and the synthetic FSV dataset.}
\label{tab:datasets}
\vspace{-1em}
\end{table}

As shown in \reffig{fig:dscr} we use  double back-propagation~\cite{DrLe91} to `expose' the gradient sent from a task $t$ to a shared layer $l$, say $g_{t}(l)$. By exposing the variable we mean that we unfold its computation graph, which in turn allows us to back-propagate through its computation.
By back-propagating on the gradients we can force them to be statistically indistinguishable across tasks through adversarial training. 

In particular we train a task classifier on top of the gradients lying at the interface of the task-specific and shared networks and use sign negation to make the task classifier fail \cite{GaLe15}. This amounts to solving the following optimization problem in terms of the discriminator weights, $w_{D}$ and the network weights, $w_{N}$:
\begin{equation}
\mathrm{min}_{\mathrm{w}_{D}}\mathrm{max}_{\mathrm{w}_{N}} L(D(g_t(w_N),w_D),t),
\end{equation}
where $g_t(w_N)$ is the gradient of task $t$ computed with $w_N$, $D(\cdot,w_D)$ is the discriminator's output for input $\cdot$, and $L(\cdot,t)$ is the cross-entropy loss for label $t$ that indicates the source task of the gradient.

Intuitively this forces every task to do its own processing within its own blocks, so that it does not need from the shared network anything different from the other tasks. This results in a separation of the network into disentangled task-specific and shared compartments.

\begin{table}[t]
	\centering
	\rowcolors{2}{white}{white}
	\resizebox{.9\linewidth}{!}{%
		\begin{tabular}{lc|ccc}
			Task        & Dataset       & Metric    & R-101    & strong baseline\\
			\shline
			Edge        & BSDS500       & odsF~$\uparrow$      & 82.5      & 81.3~\cite{Kokk16}\\
			S.Seg       & VOC           & mIoU~$\uparrow$      & 78.9      & 79.4~\cite{Che+18}\\
			H. Parts    & P. Context    & mIoU~$\uparrow$      & 64.3      & 64.9*~\cite{Che+17}\\
			Normals     & NYUD          & mErr~$\downarrow$    & 20.1      & 19.0~\cite{Ban+17}\\
			Saliency    & PASCAL-S      & maxF~$\uparrow$      & 84.0      & 83.5~\cite{Kokk17}\\
			Depth       & NYUD          & RMSE~$\downarrow$    & 0.56      & 0.58~\cite{Xu+18}\\
		\end{tabular}%
	}
	\vspace{2mm}
	\caption{\textbf{Architecture capacity}: We report upper-bounds of performance that can be reached on various competitive (but inhomogeneous)  datasets by our architecture, and compare to  strong task-specific baselines. All experiments are initialized from ImageNet pre-trained weights ($^*$ means that COCO pre-training is included). The arrow indicates better performance for each metric.}
\label{tab:ablation:sanity_check}
\end{table}

\section{Experimental Evaluation}
\label{sec:experiments}

\textbf{Datasets}
We validate our approach on different datasets and tasks. We focus on dense prediction tasks that can be approached with fully convolutional architectures. Most of the experiments are carried out on the PASCAL~\cite{Eve+12} benchmark, which is popular for dense prediction tasks. We also conduct experiments on the smaller NYUD~\cite{Sil+12} dataset of indoor scenes, and the recent, large scale FSV~\cite{Krah18} dataset of synthetic images. Statistics, as well as the different tasks used for each dataset are presented in Table~\ref{tab:datasets}.

\begin{table}[t]
\subfloat[\textbf{Baselines.} Using SE blocks in ResNet backbones (SE-bb) improves results. In all our experiments we use SE-bb baselines for fair comparison.]{
\label{tab:exp:pascal_ablation:imp_baselines}
\makebox[1.0\linewidth]{
\tablestyle{2.2pt}{1.05}
\begin{tabular}{cc|x{24}x{22}x{24}x{25}x{22}}
	     	  SE-bb         &  \#T & Edge $\uparrow$  & Seg $\uparrow$ & Parts $\uparrow$ & Norm $\downarrow$ & Sal $\uparrow$    \\
        \shline
	 		             &  1   &  70.3             & 63.98 	        & 55.85	           & 15.11            & 63.92     		  \\
	   	\checkmark       &  1   &  \bftab{71.3}	    & \bftab{64.93}  	& \bftab{57.12}	   & \bftab{14.90}    & \bftab{64.17}     \\
        \hline
	 		             &  5   &  68.0	    	    & 58.59		  		& 53.80            & \bftab{16.68}    & 60.71     		  \\ 
	    \checkmark       &  5   &  \bftab{69.2}		& \bftab{60.20}		& \bftab{54.10}	   & 17.04            & \bftab{62.10}     \\ 
\end{tabular}
}}\\
\subfloat[\textbf{Modulation.} Both SE and RA are effective modulation methods.]{
\label{tab:exp:pascal_ablation:modulation}
\makebox[1.0\linewidth]{
\tablestyle{2.2pt}{1.05}
\begin{tabular}{ccc|x{24}x{22}x{24}x{25}x{22}|r}
	SE     		    & RA         &  \#T & Edge $\uparrow$   & Seg $\uparrow$ & Parts $\uparrow$ & Norm $\downarrow$   & Sal $\uparrow$  & $\Delta_m\% \downarrow$\\
        \shline
	   		        &            &  1   &  71.3	 	         & 64.93	   	 & 57.12	        & 14.90   & 64.17 &      		   \\
	   		        &            &  5   &  69.2		         & 60.20	     & 54.10            & 17.04   & 62.10  & 6.62		   \\ 
        \hline
	  	   	        & \checkmark &  5   &  70.5		         & 62.80	     & 56.41            & 15.27   & 64.84 & \bftab{1.42}   \\
        \checkmark  &            &  5   &  71.1		         & 64.00	     & 56.84            & 15.05   & 64.35 & \bftab{0.59}   \\
\end{tabular}
}}\\
\subfloat[\textbf{SE modulation.} Modulating varying portions of the network (e.g. encoder or decoder) allows trading off performance and computation.]{
\label{tab:exp:pascal_ablation:se_enc_dec}
\makebox[1.0\linewidth]{
\tablestyle{2.2pt}{1.05}
\begin{tabular}{lcc|x{24}x{22}x{24}x{25}x{22}|r}
	enc     		& dec        &  \#T & Edge $\uparrow$   & Seg $\uparrow$ & Parts $\uparrow$  & Norm $\downarrow$   & Sal $\uparrow$  & $\Delta_m\% \downarrow$ \\
        \shline
	   		        &            &  1   &  71.3	 	         & 64.93	   	 & 57.12	        & 14.90   & 64.17 &      		   \\
	   		        &            &  5   &  69.2		         & 60.20	     & 54.10            & 17.04   & 62.1  & 6.62		   \\ 
        \hline
		            & \checkmark &  5   &  70.6		         & 63.33		 & 56.73      		& 15.14   & 63.23 & \bftab{1.44}   \\ 
	\checkmark 	    & \checkmark &  5   &  71.1		         & 64.00	     & 56.84            & 15.05   & 64.35 & \bftab{0.59}   \\
\end{tabular}
}}\\
\subfloat[\textbf{Adversarial training} is beneficial both w/ and w/o SE modulation.]{
\label{tab:exp:pascal_ablation:adversarial}
\makebox[1.0\linewidth]{
\tablestyle{2.2pt}{1.05}
\begin{tabular}{lcc|x{24}x{22}x{24}x{25}x{22}|r}
	mod      		& A          &  \#T & Edge $\uparrow$    & Seg $\uparrow$ & Parts $\uparrow$ & Norm $\downarrow$ & Sal $\uparrow$  & $\Delta_m\% \downarrow$\\
        \shline
	   		        &            &  1   &  71.3	 	         & 64.93	   	 & 57.12	        & 14.90   		 & 64.17  &       		  \\
        \hline

	   		        &            &  5   &  69.2		         & 60.20	     & 54.10            & 17.04   		 & 62.10  & 6.62		  \\
		            & \checkmark &  5   &  69.7              & 62.20	     & 55.04	        & 16.17          & 62.19  & \bftab{4.34}  \\
        \hline
	\checkmark  	&            &  5   &  71.1		         & 64.00	     & 56.84            & 15.05   		 & 64.35  & \bftab{0.59}  \\
	\checkmark      & \checkmark &  5   &  71.0		         & 64.61		 & 57.25            & 15.00          & 64.70  & \bftab{0.11}   \\
\end{tabular}
}}\\
\subfloat[\textbf{Backbones.} Improvements from SE modulation with adversarial training (SEA) are observed regardless of the capacity/depth of the backbones.]{
\label{tab:exp:pascal_ablation:backbones}
\makebox[1.0\linewidth]{
\tablestyle{2.2pt}{1.05}
\begin{tabular}{lcc|x{24}x{22}x{24}x{25}x{22}|r}
	backbone & SEA           & \#T   & Edge $\uparrow$	& Seg $\uparrow$ & Parts $\uparrow$		& Norm $\downarrow$		& Sal $\uparrow$  & $\Delta_m\% \downarrow$ \\
        \shline
	R-26 	 &               &  1    &  71.3	& 64.93	   & 57.12	    & 14.90     & 64.17    &                \\ 
	R-26     &               &  5	 &  69.2	& 60.20    & 54.10      & 17.04   	& 62.10    & 6.62           \\ 
	R-26     &  \checkmark   &  5    &  71.0	& 64.61	   & 57.25      & 15.00     & 64.70    & \bftab{0.11}   \\ 
        \hline
	R-50	 &               &  1	 &  72.7	& 68.30   & 60.70		& 14.61		& 65.40    &                \\ 
	R-50	 &               &  5	 &  69.2    & 63.20   & 55.10		& 16.04	    & 63.60    & 6.81           \\ 
	R-50     &  \checkmark   &  5	 &  72.4	& 68.00   & 61.12		& 14.68		& 65.71    & \bftab{0.04}   \\ 
        \hline
	R-101	 &               &  1	 &  73.5	& 69.76	   & 63.48		& 14.15		& 67.41    &                \\ 
	R-101	 &               &  5    &  70.5	& 66.45	   & 61.54		& 15.44		& 66.39    & 4.50           \\ 
	R-101    &  \checkmark   &  5	 &  73.5	& 68.51	   & 63.41		& 14.37		& 67.72    & \bftab{0.60}   \\ 
\end{tabular}%
}}
\vspace{.5em}
\caption{\textbf{Ablations on PASCAL}. We report average relative performance drop ($\Delta_m\%$) with respect to single task baselines. Backbone is R-26 unless otherwise noted.}
\label{tab:exp:pascal_ablations}
\vspace{-3mm}
\end{table}

\begin{table}[t]
\subfloat[Results on \textbf{NYUD-v2}.]{
\makebox[1.0\linewidth]{
\tablestyle{2.4pt}{1.1}
\begin{tabular}{cc|x{28}x{28}x{28}x{28}|c}
	 SEA          & \#T	& Edge $\uparrow$ & Seg $\uparrow$ & Norm $\downarrow$    & Depth  $\downarrow$  & $\Delta_m\% \downarrow$ \\
     \shline
                  & 1	&  74.4           & 32.82	       & 23.30                & 0.61                 &                    \\
	              & 4	&  73.2		      & 30.95	       & 23.34                & 0.70                 &  5.44              \\
	 \checkmark   & 4   &  74.5		      & 32.16	       & 23.18                & 0.57                 &  \bftab{-1.22}     \\ 
\end{tabular}%
}}\\
\subfloat[Results on \textbf{FSV}.]{
\makebox[1.0\linewidth]{
\tablestyle{2.4pt}{1.1}
\begin{tabular}{cc|x{30}x{30}x{30}|c}
	  SEA          &   \#T	     & Seg $\uparrow$ & Albedo $\downarrow$ & Disp $\downarrow$ & $\Delta_m\% \downarrow$ \\
      \shline
		           &	1        & 71.2		      & 0.086	            & 0.063             &                   \\ 
	               &    3        & 66.9		      & 0.093	            & 0.078             & 7.04              \\
      \checkmark   &    3        & 70.7		      & 0.085	            & 0.063             & \bftab{-0.02}     \\ 
\end{tabular}%
}}
\vspace{.3em}
\caption{Improvements from SE with modulation (SEA) transfer to \textbf{NYUD-v2 and FSV} datasets. We report average performance drop with respect to single task baselines. We use R-50 backbone.}
\label{tab:exp:results:nyud_fsv}
\vspace{-5mm}
\end{table}

\begin{figure*}[t]
\centering
\resizebox{1\linewidth}{!}{%

  \includegraphics[width=.9\linewidth]{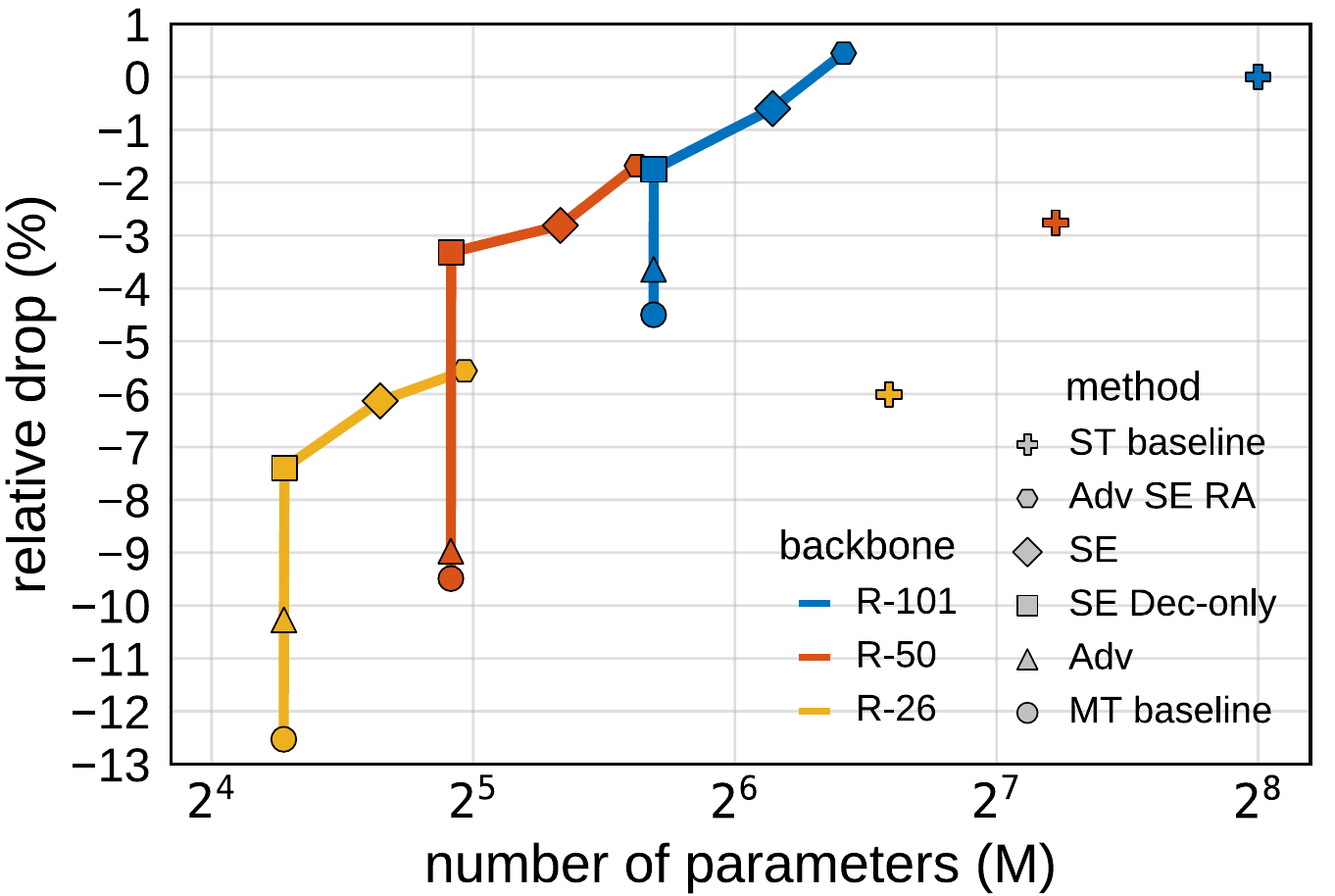}\hspace{15mm}
  \includegraphics[width=.9\linewidth]{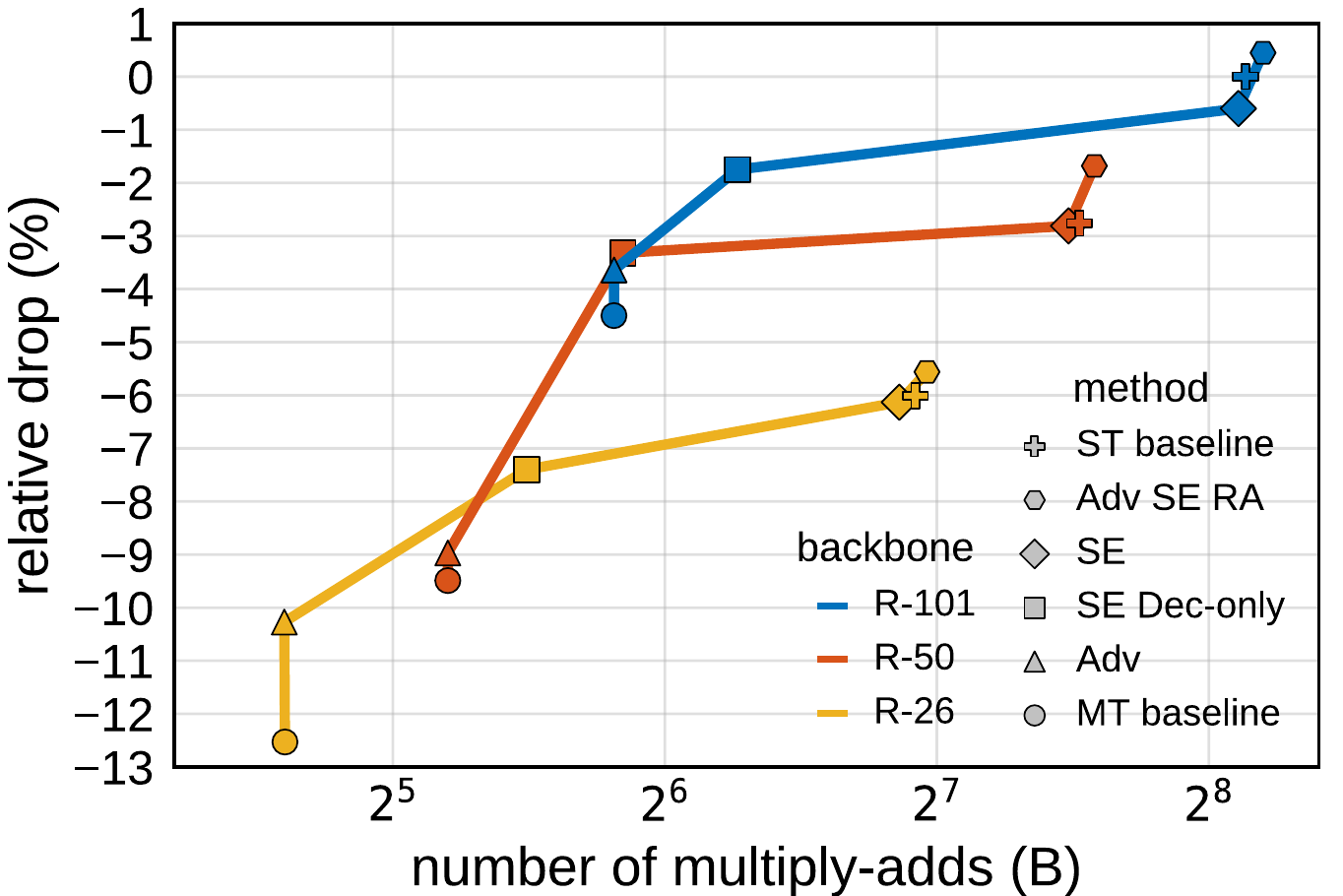}
  }
\caption{\textbf{Performance vs. Resources:} Average relative drop ($\Delta_m\%$) as a function of the number of parameters (left), and multiply-adds (right), for various points of operation of our method. We compare 3 different backbone architectures, indicated with different colors. We compare against single-tasking baseline (ST baseline), and multi-tasking baseline (MT baseline). Performance is measured relative to the best single-tasking model (R-101 backbone). An increase in performance comes for free with adversarial training (Adv). Modulation per task (SE) results in large improvements in performance, thanks to the disentangled graph representations, albeit with an increase in computational cost if used throughout the network, instead of only on the decoder (SE Dec-only vs. SE). We observe a drastic drop in number of parameters needed for our model in order to reach the performance of the baseline (SE, Adv). By using both modulation and adversarial training (Adv SE RA), we are able to reach single-task performance, with far fewer parameters.}
\label{fig:resources}
\vspace{-.5em}
\end{figure*}

\textbf{Base architecture:}
We use our re-implementation of Deeplab-v3+~\cite{Che+18} as the base architecture of our method, due to its success on dense semantic tasks. Its architecture is based on a strong ResNet encoder, with a-trous convolutions to preserve reasonable spatial dimensions for dense prediction. We use the latest version that is enhanced with a parallel a-trous pyramid classifier (ASPP) and a powerful decoder. We refer the reader to~\cite{Che+18} for more details. The ResNet-101 backbone used in the original work is replaced with its Squeeze-and-Excitation counterpart (Fig.~\ref{fig:arch_se}), pre-trained on ImageNet~\cite{Rus+15}. The pre-trained SE modules serve as an initialization point for the task-specific modulators for multi-tasking experiments.

The architecture is tested for a single task in various competitive benchmarks for dense prediction: edge detection, semantic segmentation, human part segmentation, surface normal estimation, saliency, and monocular depth estimation. We compare the results obtained with various competitive architectures. For edge detection we use the BSDS500~\cite{Mar+01} benchmark and its optimal dataset F-measure (\texttt{odsF})~\cite{MFM04}. For semantic segmentation we train on PASCAL VOC \texttt{trainaug}~\cite{Eve+12, Har+11} (10582 images), and evaluate on the validation set of PASCAL using mean intersection over union (\texttt{mIoU}). For human part segmentation we use PASCAL-Context~\cite{Che+14} and \texttt{mIoU}. For surface normals we train on the raw data of NYUD~\cite{Sil+12} and evaluate on the test set using mean error (\texttt{mErr}) in the predicted angles as the evaluation metric. For saliency we follow~\cite{Kokk17} by training on MSRA-10K~\cite{Che+15}, testing on PASCAL-S~\cite{Li+14} and using the maximal F-measure (\texttt{maxF}) metric. Finally, for depth estimation we train and test on the fully annotated training set of NYUD using root mean squared error (\texttt{RMSE}) as the evaluation metric. For implementation details, and  hyper-parameters, please refer to the Appendix.

Table~\ref{tab:ablation:sanity_check} benchmarks our architecture against popular state-of-the-art methods. We obtain competitive results, for all tasks. We emphasize that these benchmarks are inhomogeneous, i.e. their images are not annotated with all tasks, while including domain shifts when training for multi-tasking (eg. NYUD contains only indoor images). In order to isolate performance gains/drops as a result of multi-task learning (and not domain adaptation, or catastrophic forgetting), in the experiments that follow, we use homogeneous datasets.

\begin{figure*}[t]
\begin{center}
\resizebox{1\linewidth}{!}{%
\includegraphics[width=0.95\linewidth]{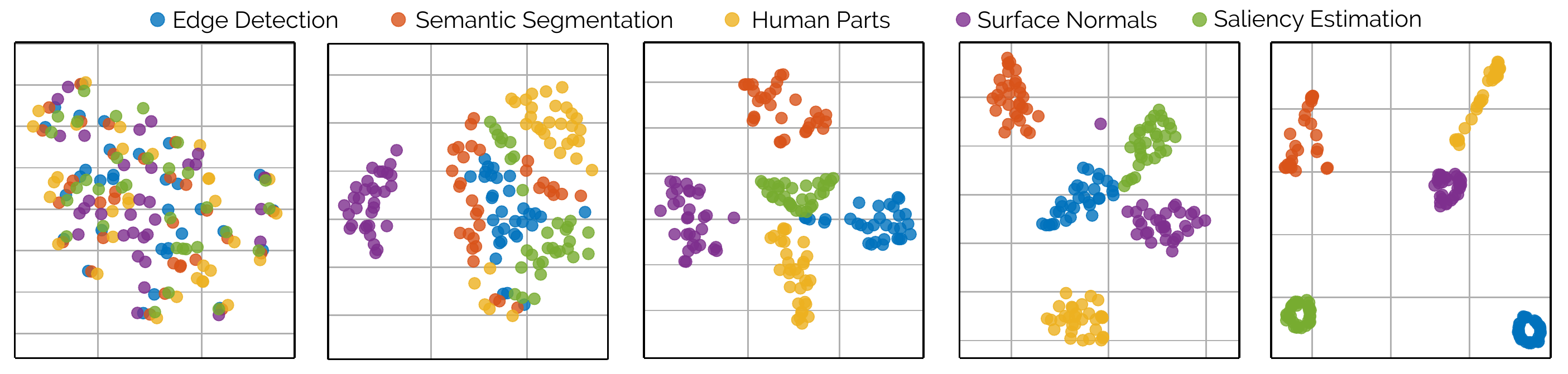}
}
\end{center}
\vspace{-1em}
\caption{\textbf{t-SNE visualization of task-dependent feature activations of a single image} at increasing depths of the network (from left to right). Features in early layers are more similar across tasks and progressively get more adapted to specific tasks in later layers.}
\label{fig:se_ws_tsne}
\vspace{-.5em}
\end{figure*}

\textbf{Multi-task learning setup:}
We proceed to multi-tasking experiments on PASCAL. We keep the splits of PASCAL-Context, which provides labels for edge detection, semantic segmentation, and human part segmentation. In order to keep the dataset homogeneous and the architecture identical for all tasks, we did not use instance level tasks (detection, pose estimation) that are provided with the dataset. To increase the number of tasks we automatically obtained ground-truth for surface normals and saliency through label-distillation using pre-trained state-of-the-art models (\cite{Ban+17} and~\cite{Che+18}, respectively), since PASCAL is not annotated with those tasks. For surface normals, we masked out predictions from unknown and/or invalid classes (eg. sky) during both training and testing. In short, our benchmark consists of 5 diverse tasks, ranging from low-level (edge detection, surface normals), to mid-level (saliency) and high-level (semantic segmentation, human part segmentation) tasks.

\textbf{Evaluation metric:} We compute multi-tasking performance of method $m$ as the average per-task drop with respect to the single-tasking baseline $b$ (i.e different networks trained for a single task each):
\begin{equation}
\Delta_m = \frac{1}{T} \sum _{i=1}^T (-1)^{l_i}\left(M_{m,i}-M_{b,i}\right)/M_{b,i}
\end{equation}
where $l_i=1$ if a lower value means better for measure $M_i$ of task $i$, and $0$ otherwise. Average relative drop is computed against the baseline that uses {\emph{ the same backbone}}.

To better understand the effect of different aspects of our method, we conduct a number of ablation studies and present the results in Table~\ref{tab:exp:pascal_ablations}.

We construct a second baseline, which tries to learn all tasks simultaneously with a single network, by connecting $T$ task-specific convolutional classifiers ($1\times1$ conv layers) at the end of the network. As also reported by~\cite{Kokk17}, a non-negligible average performance drop can be observed (-6.6\% per task for R-26 with SE). We argue that this drop is mainly triggered by conflicting gradients during learning.

\textbf{Effects of modulation and adversarial training:}
Next, we introduce the modulation layers described in Section~\ref{sec:modulation}. We compare parallel residual adapters to SE (Table~\ref{tab:exp:pascal_ablation:modulation}) when used for task modulation. Performance per task recovers immediately by separating the computation used by each task during learning (-1.4 and -0.6 vs. -6.6 for adapters and SE, respectively). SE modulation results in better performance, while using slightly fewer parameters per task. We train a second variant where we keep the computation graph identical for all tasks in the encoder, while using SE modulation only in the decoder (Table~\ref{tab:exp:pascal_ablation:se_enc_dec}). Interestingly, this variant reaches the performance of residual adapters (-1.4), while being much more efficient in terms of number of parameters and computation, as only one forward pass of the encoder is necessary for all tasks.

In a separate experiment, we study the effects of adversarial training described in Section~\ref{sec:discriminator}. We use a simple, fully convolutional discriminator to classify the source of the gradients. Results in Table~\ref{tab:exp:pascal_ablation:adversarial} show that adversarial training is beneficial for multi-tasking, increasing performance compared to standard multi-tasking (-4.4 vs -6.6). Even though the improvements are less significant compared to modulation, they come without extra parameters or computational cost, since the discriminator is used only during training.

The combination of SE modulation with adversarial training (Table~\ref{tab:exp:pascal_ablation:adversarial}) leads to additional improvements (-0.1\% worse than the single-task baseline), while further adding residual adapters surpasses single-tasking (+0.45\%), at the cost of 12.3\% more parameters per task (Fig.~\ref{fig:resources}).

\textbf{Deeper Architectures:} Table~\ref{tab:exp:pascal_ablation:backbones} shows how modulation and adversarial training perform when coupled with deeper architectures (R-50 and R-101). The results show that our method is invariant to the depth of the backbone, consistently improving the standard multi-tasking results.

\begin{figure*}[h]
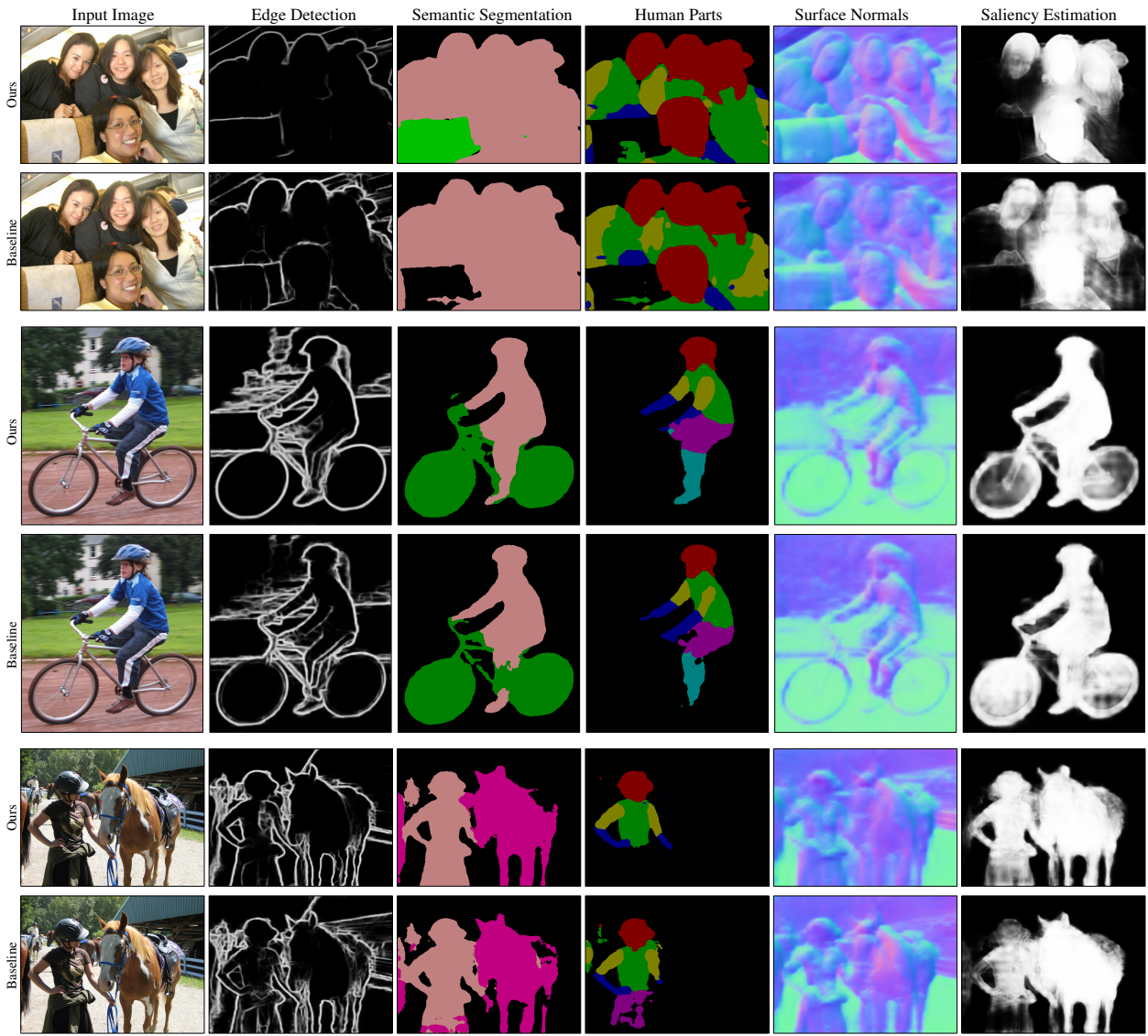

	\centering
	{\scriptsize \hspace{5mm} Input Image \qquad \qquad \hspace{3mm} Edge Detection \qquad \qquad \hspace{-3mm} Semantic Segmentation \qquad \qquad \hspace{-2mm} Human Parts \qquad \qquad Surface Normals \qquad \qquad Saliency Estimation}
	\showonerowpascal{2008_000277}{qualitative}{0.5}{Ours}{35}{.95}{\Huge}
	\showonerowpascal{2008_000277_baseline}{qualitative}{0.5}{Baseline}{28}{.95}{\Huge} \vspace{1mm}
	\showonerowpascal{2009_000461}{qualitative}{0.5}{Ours}{35}{.95}{\LARGE}
	\showonerowpascal{2009_000461_baseline}{qualitative}{0.5}{Baseline}{28}{.95}{\LARGE} \vspace{1mm}
	\showonerowpascal{2010_004542}{qualitative}{0.5}{Ours}{35}{.95}{\Huge}
	\showonerowpascal{2010_004542_baseline}{qualitative}{0.5}{Baseline}{28}{.95}{\Huge} \vspace{1mm}

	\caption{\textbf{Qualitative Results on PASCAL}: We compare our model against standard multi-tasking. For the baseline, features from edge detection appear in saliency estimation results (second row), and surface normals are noisy, indicating the need to disentangle the learned representations.}
	\label{fig:qualitative_pascal}
	\vspace{1mm}
\end{figure*}

\begin{figure*}[h]
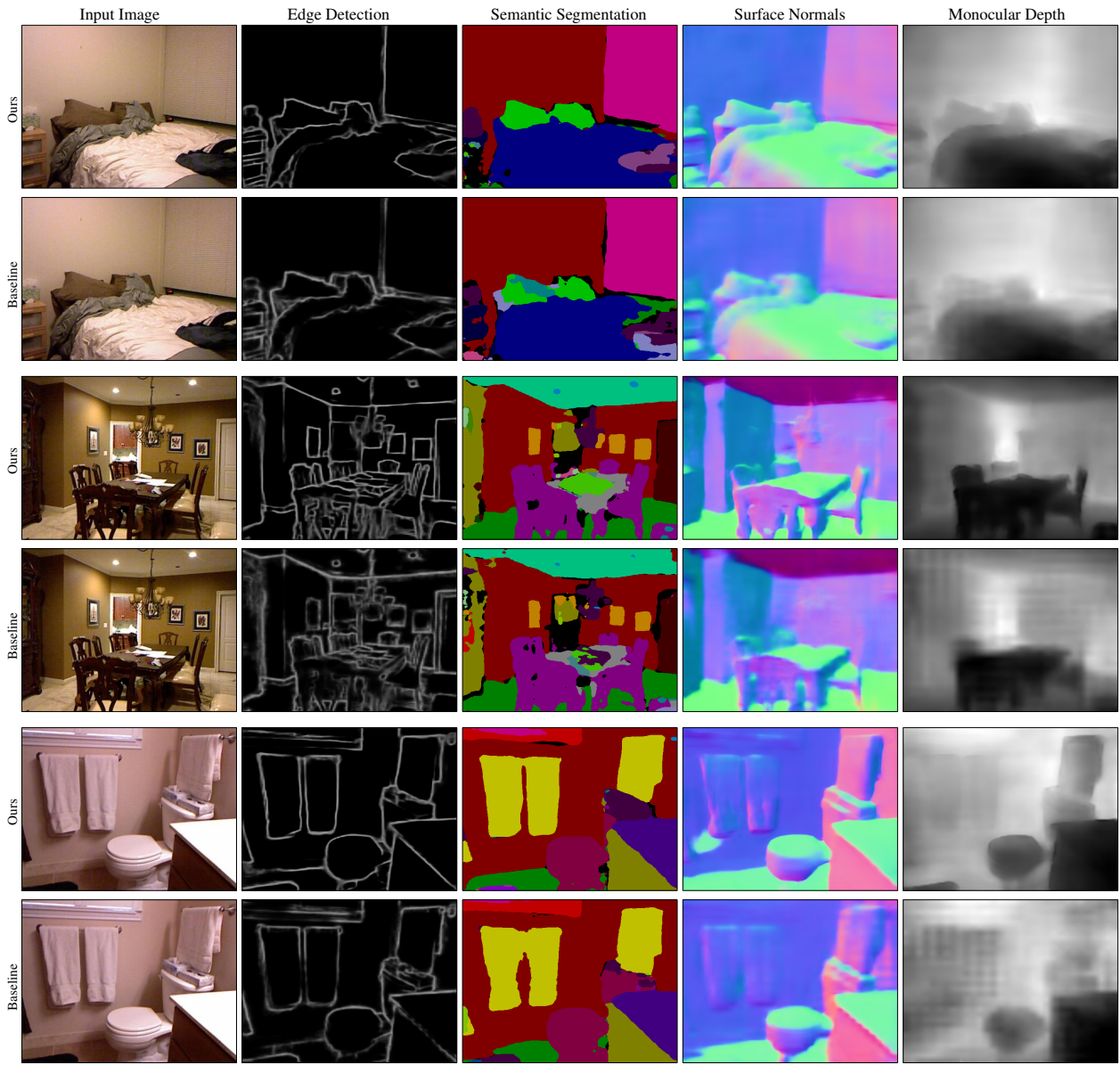

	\centering
	{\scriptsize \hspace{2mm} Input Image \qquad \qquad \hspace{8mm} Edge Detection \qquad \qquad \hspace{4mm} Semantic Segmentation \qquad \qquad \hspace{2mm} Surface Normals \qquad \qquad \qquad Monocular Depth}
    \showonerownyud{00056}{qualitative}{0.5}{Ours}{35}{.95}
	\showonerownyud{00056_baseline}{qualitative}{0.5}{Baseline}{28}{.95} \vspace{1mm}
	\showonerownyud{01397}{qualitative}{0.5}{Ours}{35}{.95}
	\showonerownyud{01397_baseline}{qualitative}{0.5}{Baseline}{28}{.95} \vspace{1mm}
	\showonerownyud{00694}{qualitative}{0.5}{Ours}{35}{.95}
	\showonerownyud{00694_baseline}{qualitative}{0.5}{Baseline}{28}{.95} \vspace{1mm}
	
	\caption{\textbf{Qualitative Results on NYUD}: We compare our model against standard multi-tasking. The baseline predicts blurry edges and depth, as well as inconsistent labels on the pillow (where surface normals change). Our method is able to recover from these issues.}
	\label{fig:qualitative_nyud}
	\vspace{-2mm}
\end{figure*}

\textbf{Resource Analysis:} Figure~\ref{fig:resources} illustrates the performance of each variant as a function of the number of parameters, as well as the FLOPS (multiply-adds) used during inference. We plot the relative average per-task performance compared to the single-tasking R-101 variant (blue cross), for the 5 tasks of PASCAL. Different colors indicate different backbone architectures. We see a clear drop in performance by standard multi-tasking (crosses vs. circles), but with fewer parameters and FLOPS. Improvements due to adversarial training come free of cost (triangles) with only a small overhead for the discriminator during training.

Including modulation comes with significant improvements, but also with a very slight increase of parameters and a slight increase of computational cost when including the modules on the decoder (rectangles). The increase becomes more apparent when including those modules in the encoder as well (diamonds). Our most accurate variant using all of the above (hexagons) outperforms the single-tasking baselines by using only a fraction of their parameters.

We note that the memory and computational complexities of the SE blocks and the adapters are negligible, but since it affects the outputs of the layer it means that we cannot share the computation of the ensuing layers across all tasks, and thus the increased number of multiply-adds.

\textbf{Learned Disentangled Representations:} In order to highlight the importance of task modulation, we plot the learned representations for different tasks in various depths of our network. Figure~\ref{fig:se_ws_tsne} shows the t-SNE representations~\cite{MaHi08} of the SE activations in equally spaced levels of the network. The activations are averaged for the first 32 samples of the validation set, following~\cite{HSS18}, and they are sorted per task. The resulting plots show that in the early stages of the network the learned representations are almost identical. They gradually become increasingly different as depth grows, until they are completely different for different tasks at the level of the classifier. We argue that this disentanglement of learned representations also translates to performance gains, as shown in Table~\ref{tab:exp:pascal_ablations}.

\textbf{Validation on additional datasets:} We validate our approach in two additional datasets, NYUD~\cite{Sil+12} and FSV~\cite{Krah18}.
NYUD is an indoor dataset, annotated with labels for instance edge detection, semantic segmentation into 41 classes, surface normals, and depth.
FSV is a large-scale synthetic dataset, labelled with semantic segmentation (3 classes), albedo, and depth (disparity).

Table~\ref{tab:exp:results:nyud_fsv} presents our findings for both datasets. As in PASCAL, when we try to learn all tasks together, we observe a non-negligible drop compared to the single-tasking baseline. Performance recovers when we plug in modulation and adversarial training. Interestingly, in NYUD and FSV we observe larger improvements compared to PASCAL. Our findings are consistent with related works~\cite{Xu+18,EiFe15} which report improved results for multi-tasking when using depth and semantics. %

Figures~\ref{fig:qualitative_pascal} and~\ref{fig:qualitative_nyud} illustrate some qualitative examples, obtained by our method on PASCAL and NYUD, respectively. Results in each row are obtained with a single network. We compare our best model to the baseline architecture for multi-tasking (without per-task modulation, or adversarial training). We observe a quality degradation in the results of the baseline. Interestingly, some errors are obtained clearly as a result of standard multi-tasking. Edge features appear during saliency estimation in Fig~\ref{fig:qualitative_pascal}, and predicted semantic labels change on the pillows, in areas where the surface normals change, in Fig~\ref{fig:qualitative_nyud}. In contrast, our method provides disentangled predictions that are able to recover from such issues, reach, and even surpass the single-tasking baselines.

\section{Conclusions}

In this work we have shown that we can attain, and even surpass single-task performance through multi-task networks, provided we execute one task at a time. We have achieved this by introducing a method that allows a network to `focus' on the task at hand in terms of task-specific feature modulation and adaptation. 

In a general vision architecture one can think of task attention as being determined  based on the operation currently being performed - e.g. using object detection to find an object, normal estimation and segmentation to grasp it. Tasks can also be executed in an interleaved manner, with low-level tasks interacting with high-level ones in a bottom-up/top-down cascade \cite{mumford}. We intend to explore these directions in future research. 

\section*{Appendix}
The following additional information is provided in the appendix:

\begin{itemize}
\item Ablated results using different backbone architectures for NYUD and FSV datasets in Appendix A.
\item Baselines using UberNet-type networks with and without the proposed modulation and adversarial training in Appendix B.
\item Results using MobileNet as the backbone architecture, in order to highlight potential mobile phone applications in Appendix C.
\item Technical details for training and testing in Appendix D. Code will also be published.
\end{itemize}

\section*{Appendix A: More results on NYUD and FSV}
\label{sec:nyud_fsv}
Table~\ref{tab:nyud_fsv} illustrates the quantitative results obtained by our method on NYUD~\cite{Sil+12} and FSV~\cite{Krah18}, by changing the backbone architecture. Results are consistent among backbones, and by including modulation and adversarial training to the pipeline, we get improved results with respect to the multi-task and single-task baselines, irrespective of the network depth.

\begin{table}[t]

\subfloat[Our method on NYUD~\cite{Sil+12}, for different backbones.]{
\label{tab:nyud_fsv:nyud}
\makebox[1\linewidth]{
\tablestyle{2.2pt}{1.05}
\begin{tabular}{lcc|x{24}x{22}x{25}x{27}|r}
	backbone & SEA           & \#T   & Edge $\uparrow$	& Seg $\uparrow$& Norm $\downarrow$	& Depth $\downarrow$   & $\Delta_m\% \downarrow$ \\
        \shline
	R-26	  &               &  1	 &  72.9	        & 29.87     	& 24.34	            & 0.650    		       & 0                 \\ 
	R-26	  &               &  4	 &  72.4	        & 27.74	        & 24.83	            & 0.729    		       & 5.50              \\ 
	R-26      &  \checkmark   &  4	 &  73.5	        & 30.07	        & 24.316	        & 0.625     		   & \bftab{-1.36}     \\
	    \hline
    R-50	  &               & 1	&  74.4             & 32.82	        & 23.3              & 0.610                &                   \\
	R-50	  &               & 4	&  73.2		        & 30.95	        & 23.34             & 0.700                &  5.44             \\
	R-50      &  \checkmark   & 4   &  74.5		        & 32.16	        & 23.18             & 0.570                &  \bftab{-1.22}    \\ 
	    \hline
    R-101	  &               &  1	 &  74.9	        & 35.90        & 22.90	            & 0.580                &                   \\
	R-101	  &               &  4	 &  73.8	        & 31.20        & 23.07	            & 0.650                &  6.63             \\
	R-101     &  \checkmark   &  4   &  75.6	        & 35.60        & 22.73	            & 0.560                &  \bftab{-1.07}    \\ 
\end{tabular}%
}}\\

\subfloat[Our method on FSV~\cite{Krah18}, for different backbones.]{
\label{tab:nyud_fsv:fsv}
\makebox[1\linewidth]{
\tablestyle{2.2pt}{1.05}
\begin{tabular}{lcc|x{22}x{30}x{27}|r}
	backbone & SEA           & \#T 	& Seg $\uparrow$    & Albedo $\downarrow$	& Depth $\downarrow$ & $\Delta_m\% \downarrow$\\
        \shline
    R-26	  &               &  1	 & 69.77	        & 0.087 	        & 0.065             & 0                  \\ 
	R-26	  &               &  3	 & 66.71	        & 0.090             & 0.073             & 6.41	             \\
	R-26      &  \checkmark   &  3	 & 71.36	        & 0.085	            & 0.065             & \bftab{-1.80}      \\
	    \hline
	R-50	  &               &  1	 & 71.14		    & 0.086	            & 0.063             & 0                  \\ 
	R-50	  &               &  3	 & 66.90	        & 0.093	            & 0.078             & 7.04	              \\
	R-50      &  \checkmark   &  3	 & 70.69		    & 0.085	            & 0.063             & \bftab{-0.02}      \\
	    \hline
	R-101	  &               &  1	 & 72.10	        & 0.086     	    & 0.063             & 0                  \\ 
	R-101	  &               &  3	 & 68.12	        & 0.091	            & 0.072             & 8.75	              \\
	R-101     &  \checkmark   &  3	 & 72.24	        & 0.083	            & 0.062             & \bftab{-1.57}      \\
\end{tabular}%
}}\\

\caption{\textbf{Our method for NYUD, and FSV}: Results with different backbones: R-26, R-50, and R-101. Negative drop indicates improved performance with respect to the single-tasking baseline. Arrows indicate desired behaviour of each metric.}

\label{tab:nyud_fsv}
\end{table}

\section*{Appendix B: UberNet-type baseline}
\label{sec:ubernet}

The architecture of~\cite{Kokk17} learns multiple tasks by using a common backbone and a light-weight decoder (skip connections and $1\times1$ convolutions) per task. We re-implement the UberNet architecture, and we substitute the VGG~\cite{SiZi15} backbone of the original work with the SE-ResNet~\cite{HSS18} used in our work. The main difference with our architecture are the skip connections and $1\times1$ convolutions that comprise each task-specific head, instead of the powerful Deeplab-v3+ ASPP decoder~\cite{Che+18} used throughout our work. Similarly to our work, and similarly to the observation of the original author, we observe a non-trivial drop in performance when learning to multi-task with a common, entirely shared backbone (Table~\ref{tab:exp:ubernet}). We plug-in SE and adversarial training and we recover most of the drop. We provide results for all 3 datasets that have been used throughout this work. Results obtained by our architecture are presented in the last row of each table. The relatively lower performance especially for semantic tasks compared to our method is due to the absence of a strong decoder. The observations regarding modulation and adversarial training are, nevertheless, consistent.

\begin{table}[t]

\subfloat[UberNet results in PASCAL. ]{
\label{tab:exp:ubernet:pascal}
\makebox[1\linewidth]{
\tablestyle{2.2pt}{1.05}
\begin{tabular}{lcc|x{24}x{22}x{24}x{25}x{22}|r}
	backbone & SEA           & \#T   & Edge $\uparrow$	& Seg $\uparrow$ & Parts $\uparrow$		& Norm $\downarrow$		& Sal $\uparrow$  & $\Delta_m\% \downarrow$ \\
        \shline
	R-50-Uber	 &               &  1	 & 71.7			& 66.90		   & 59.80		& 15.00			& 64.56     &                \\ 
	R-50-Uber	 &               &  5	 & 70.3			& 60.90		   & 57.00		& 16.65			& 62.15     & 7.10           \\ 
	R-50-Uber    &  \checkmark   &  5	 & 70.5		    & 65.50		   & 60.15		& 14.94			& 64.98     & \bftab{0.43}   \\ 
	\hline
	R-50         &  \checkmark   &  5	 & 72.4	    	& 68.00		   & 61.10		& 14.80			& 65.70     & n.a   \\
\end{tabular}%
}}\\

\subfloat[UberNet-type results in NYUD.]{
\label{tab:exp:ubernet:nyud}
\makebox[1\linewidth]{
\tablestyle{2.2pt}{1.05}
\begin{tabular}{lcc|x{24}x{22}x{25}x{27}|r}
	backbone & SEA           & \#T   & Edge $\uparrow$	& Seg $\uparrow$& Norm $\downarrow$	& Depth $\downarrow$    & $\Delta_m\% \downarrow$ \\
        \shline
	R-50-Uber	  &               &  1	 &  73.9		    & 32.50		    & 22.90			    & 0.669    		        & 0              \\ 
	R-50-Uber	  &               &  4	 &  72.3		    & 29.48		    & 24.16			    & 0.716    		    	& 6.00           \\ 
	R-50-Uber     &  \checkmark   &  4	 &  73.7		    & 31.19		    & 23.46			    & 0.632     			& \bftab{0.30}    \\
	\hline
	R-50          &  \checkmark   &  4	 &  74.5		    & 32.20		    & 23.20			    & 0.570      			& n.a    \\
\end{tabular}%
}}\\

\subfloat[UberNet results in FSV.]{
\label{tab:exp:ubernet:fsv}
\makebox[1\linewidth]{
\tablestyle{2.2pt}{1.05}
\begin{tabular}{lcc|x{22}x{30}x{27}|r}
	backbone & SEA           & \#T 	& Seg $\uparrow$    & Albedo $\downarrow$	& Depth $\downarrow$ & $\Delta_m\% \downarrow$ \\
        \shline
	R-50-Uber	  &               &  1	& 70.26		        & 0.092		            & 0.101		         & 0            \\ 
	R-50-Uber	  &               &  3	& 67.02		        & 0.093		            & 0.124		         & 9.50         \\ 
	R-50-Uber     &  \checkmark   &  3	& 69.45		        & 0.091		            & 0.111		  	     & \bftab{3.32}  \\
	\hline
	R-50          &  \checkmark   &  3 & 70.70		        & 0.085	                & 0.063             & n.a      \\
\end{tabular}%
}}\\

\caption{\textbf{UberNet for PASCAL, NYUD, and FSV}: Standard multi-task learning results in a significant drop in performance, that is recovered with modulation and adversarial training. Last rows of each table present results obtained by our architecture.}

\label{tab:exp:ubernet}
\end{table}

\begin{figure*}[t]
	\centering
	\resizebox{.8\linewidth}{!}{%
		\includegraphics[width=1.0\linewidth]{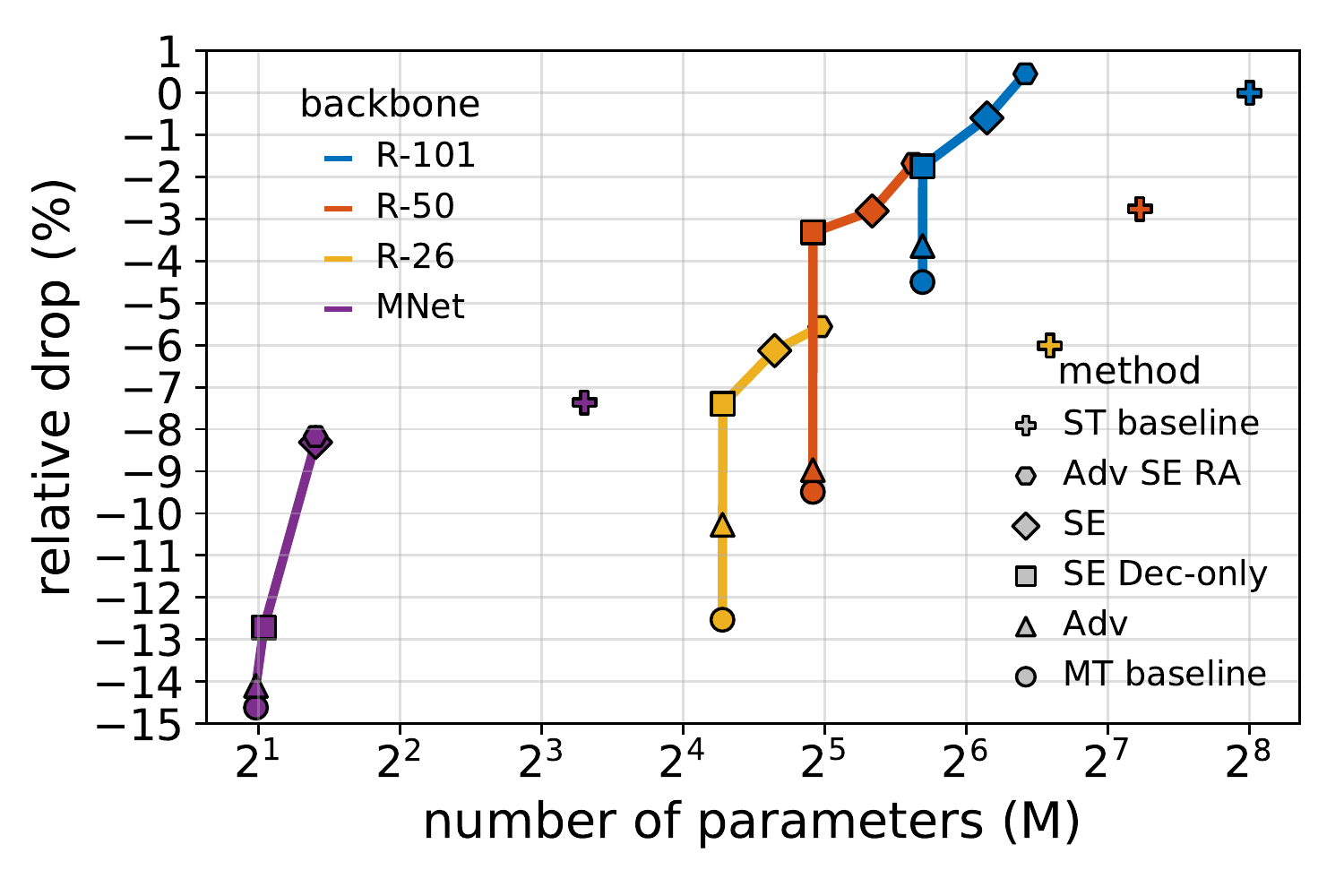}\hspace{10mm}
		\label{fig:resources_mnet:params}
		\includegraphics[width=1.0\linewidth]{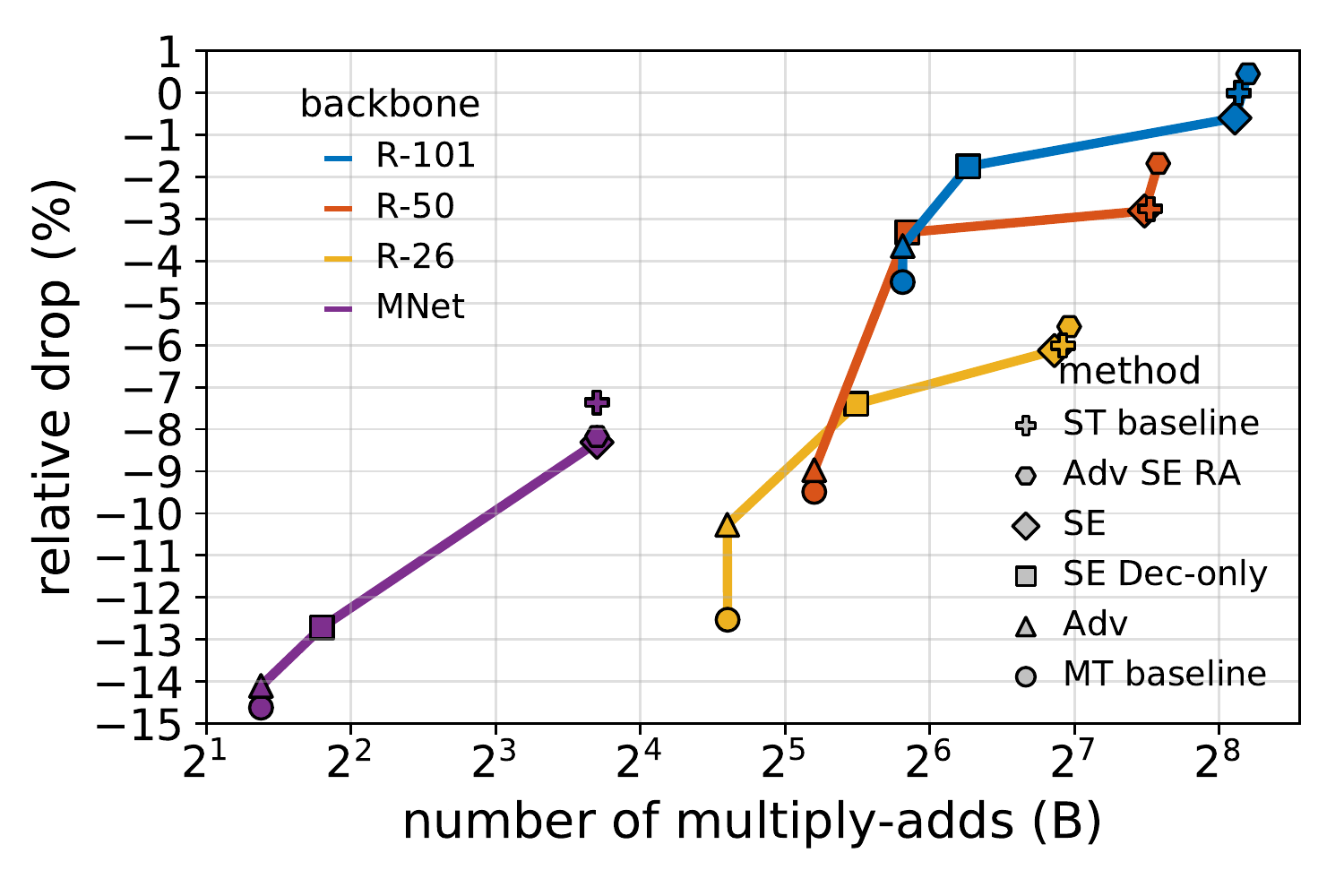}
		\label{fig:resources_mnet:flops}
	}
	
	\caption{\textbf{Performance vs. Resources for MobileNet:} Average relative drop as a function of the number of parameters (left), and multiply-adds (right), for various points of operation of our method. Different backbones are indicated with different colors. MobileNet with our modulation is able to reach R-50 results for standard multi-tasking, by using much less parameters and computation.}
	\label{fig:resources_mnet}
	\vspace{-.5em}
\end{figure*}

\section*{Appendix C: MobileNet-v2 backbone for multiple tasks}
\label{sec:mobilenet}
Our multi-tasking framework could find application in light-weight CNNs designed for mobile phone applications. For example, by using our framework, many different tasks can be executed with only a single and small set of parameters being shipped to the end user. To test this idea, we change our backbone to the light-weight MobileNet-v2~\cite{San+18}, an architecture specifically designed for mobile phones. We change the decoder accordingly: convolutions are changed to depth-wise convolutions, and ReLU activations are changed to ReLU6. We pre-train a variant that uses Squeeze and Excitation modules on ImageNet (SE-MobileNet), and fine-tune for multi-task learning. We test standard multi-tasking against the variants of our method that use SE modulation. Table~\ref{tab:mobilenet} summarizes our findings. Similarly to the experiments using SE-ResNet, disentangling the representations for each task also helps for MobileNet. By using SE per task both on the encoder and decoder, our method outperforms the single-tasking baseline. Figure~\ref{fig:resources_mnet} puts these results in perspective, comparing them to results obtained by SE-ResNet architecture. It is remarkable that by using modulation, MobileNet is no worse than the R-50 standard multi-tasking baseline using much less computational cost, and only 8\% of its parameters.


\begin{table}[t]
\centering
\resizebox{\linewidth}{!}{%
\tablestyle{2.2pt}{1.05}
\begin{tabular}{lccc|x{24}x{24}x{24}x{26}x{25}|x{35}}
	backbone  & enc        & dec          & \#T  & Edge $\uparrow$	& Seg $\uparrow$ & Parts $\uparrow$		& Norm $\downarrow$		& Sal $\uparrow$ & $\Delta_m\% \downarrow$ \\
        \shline
	MNet	 &             &              &  1	 & 69.5	        & 62.10	         & 54.88	                & 14.88 	            & 66.30          & 0                       \\ 
    MNet	 &             &              &  5	 & 67.2	        & 54.10	         & 53.00             		& 16.76	                & 62.70          & 7.57                    \\
    MNet	 &             & \checkmark   &  5	 & 67.5	        & 57.40	         & 54.50             		& 16.55	                & 63.00          & 5.47                    \\ 
	MNet   &  \checkmark & \checkmark   &  5	 & 69.2	        & 61.60	         & 55.17	                & 15.21	                & 65.60          & \bftab{0.97}            \\ 
\end{tabular}%
}
\vspace{0.1em}
\caption{\textbf{Results using MobileNet in PASCAL}: Modulation with SE is able to recover the performance that is lost using standard multi-task learning.}
\label{tab:mobilenet}
\vspace{-0.5em}
\end{table}

%
%
%

\section*{Appendix D: Implementation Details}
\label{sec:details}
In this section we provide the technical details for our implementation.

\textbf{Generic hyper-parameters:}
The entire hyper-parameter search was performed on the single-task baselines. For all tasks, we use synchronized SGD with momentum of 0.9 and weight decay of 1e-4. We set the initial learning rate to 0.005 and use the poly learning rate~\cite{Che+17}. All our models are trained on a single GPU with batch size 8, and spatial input of $512 \times 512$. We used multi-GPU training with batch size 16 and synchronous batchnorm layers only for the sanity check experiments (Table 2 of main paper), for a fair comparison with competing methods. Standard flipping, rotations, and scaling was used for data augmentation. The number of total epochs is set to 60 for PASCAL~\cite{Eve+12}, 200 for NYUD~\cite{Sil+12}, and 3 for the large-scale FSV~\cite{Krah18}.

\textbf{Weighting of the losses:}
Related work deals with automatically weighting of the losses for multi-task learning~\cite{Che+17a,SeKo18}. We compared these methods to selecting the optimal weights by grid-search. In our setup, we found out that grid-search works better, probably because of the very imbalanced optimal parameters (optimal weighting of loss for edge detection is 50 times higher than semantic segmentation, for example). In particular, we re-implemented~\cite{SeKo18} that uses multi-objective optimization to re-weight the gradients from each task, in order to optimize the shared parts of the network towards a direction useful for all tasks. This lead to better results than uniform weights, however we obtained better results by simple grid-search.

For all our experiments, when training for multiple tasks, we divide the learning rate of the shared layers by the number of tasks ($T$), since $T$ updates are happening for the same mini-batch on the shared part of the network.

During training, we used the following formula for the weights of the losses:
\begin{equation}
    L = (1 - w_d) \cdot \sum_{t=1}^{T}{w_t\cdot L_t} + w_d \cdot L_d,
\end{equation}
where $w_t$ weights the loss $L_t$ of task $t$, and $w_d$ the loss $L_d$ of the discriminator. All losses are averaged to the number of samples that the prediction contains ($W \times H \times C \times N$).

\textbf{Edge Detection:}
For edge detection, we use $w_t=50$ and binary cross-entropy loss. As is common practice~\cite{XiTu17,Kokk16,Man+17}, the positive pixels are weighted more (0.95) than the negative ones (0.05), to account for the class imbalance. When training for a single task in BSDS (Table 2 of main paper), where there are more than a single annotators, we use the multi-instance learning (MIL) loss of~\cite{Kokk16}. No MIL is used when training in PASCAL or NYUD. We follow the common evaluation on those two datasets~\cite{Man+17} by setting the maximum allowed mis-localization of edges (\texttt{maxDist} parameter) to 0.0075 and 0.011 for PASCAL and NYUD, respectively.

\textbf{Semantic Segmentation:}
For semantic segmentation, we used $w_t=1$, and cross-entropy loss. When training on VOC trainaug~\cite{Che+18} (Table 2 of main paper), we did not finetune separately on VOC val.

\textbf{Human Part Segmentation:}
For human part segmentation, we used $w_t=2$. and cross-entropy loss. Samples that do not contain any humans did not contribute to the loss.

\textbf{Surface Normals:}
For surface normal estimation, we used $w_t=10$ and $\mathcal{L}_1$ loss with unit-vector normalization. During rotation augmentation, we carefully rotate the unit vector of the surface normals accordingly, to point to a consistent direction.

\textbf{Albedo:}
For albedo, we use $w_t=10$, and standard  $\mathcal{L}_1$ loss. We emphasize that our architecture is not optimal for this task, since the output is 4 times smaller than the input, and tiny details are not captured. Using an architecture suited for albedo is out of the scope of this work.

\textbf{Monocular Depth:}
For monocular depth estimation we use $w_t=1$, and the loss of~\cite{EiFe15}, which is a combination of $\mathcal{L}_1$ loss and a smoothness term that enforces the spatial gradients of the prediction to be consistent with the ones of the ground truth. Our experiments showed that including the smoothness term to the loss leads to better results, quantitatively and qualitatively.

\textbf{Discriminator:}
We use a fully-convolutional discriminator, which consists of two $1\times1$ conv layers and a ReLU activation. We did not observe improvements when using discriminators of larger depth. We normalize the gradient of the losses of the tasks by their norm before passing them through the discriminator. This practice makes training more stable because the norm of the gradients becomes smaller as training progresses. We use $w_d=0.1$.

\textbf{Further Technical Details:}
Our work is implemented in PyTorch~\cite{Pas+17}. During weight update PyTorch applies momentum and weight decay to all modules in the definition of a network. This behaviour is not desired When using generic and task-specific weights, since the task-specific ones are only used in the forward pass of the particular task, which leads to $T-1$ unwanted updates. This behaviour is avoided by tracing the graph of computation and updating only the weights that were used, which also translated into quantitative improvements.

{\small
	\bibliographystyle{ieee}
	\bibliography{2019_cvpr}
}
	
\end{document}